\pdfoutput=1
\documentclass[11pt]{article}

\usepackage{acl}

\usepackage{times}
\usepackage{latexsym}
\usepackage[T1]{fontenc}
\usepackage[utf8]{inputenc}
\usepackage{microtype}
\usepackage{inconsolata}

\usepackage{xcolor} 

\usepackage{float}
\usepackage{footnotehyper}
\usepackage{graphicx}
\usepackage{subcaption}

\usepackage[main=english]{babel}
\usepackage[autostyle]{csquotes}
\usepackage{microtype}
\usepackage{enumitem}
\usepackage{lipsum}
\usepackage[most]{tcolorbox}

\usepackage{tabularx}
\usepackage{booktabs}
\usepackage{multirow}
\usepackage{makecell}

\makeatletter
\newcommand{\customlabel}[2]{%
 \@bsphack\begingroup
 \def\@currentlabel{#2}%
 \label{#1}%
 \endgroup\@esphack
}
\makeatother
\newcommand{\tablelabel}[1]{\customlabel{#1}{\arabic{table}}}
\newcommand{\figurelabel}[1]{\customlabel{#1}{\arabic{figure}}}

\newcommand{\sectionlabel}[1]{\customlabel{#1}{Section \arabic{section}}}
\newcommand{\subsectionlabel}[1]{\customlabel{#1}{Section \arabic{section}.\arabic{subsection}}}

\usepackage{amsmath}
\usepackage{amsfonts}

\usepackage[disable,size=tiny]{todonotes}
\usepackage{lipsum}
\newcommand{\ak}[1]{{\color{black}#1}} 
\newcommand{\se}[1]{\textcolor{black}{#1}} 

\usepackage{xparse}
\NewDocumentCommand\de{m g}{\IfNoValueTF{#2}{\textit{#1}}{\textit{#1} (\textit{en.}:~#2)}} 
\newcommand{\Frau}[0]{\texttt{Woman}}
\newcommand{\Migrant}[0]{\texttt{Migrant}}

\title{
Fine-Grained Detection of Solidarity for Women and Migrants in \\
155 Years of German Parliamentary Debates
}

\author{
    Aida Kostikova$^1$, Benjamin Paassen$^1$, \textbf{Dominik Beese}$^2$, \\
    \textbf{Ole Pütz}$^1$, \textbf{Gregor Wiedemann}$^3$, \textbf{Steffen Eger}$^{4,5}$ \\
    $^1$ Bielefeld University, $^2$ TU Darmstadt, $^3$ Hans-Bredow-Institut, \\
    $^4$ University of Mannheim, $^5$ University of Technology Nuremberg \\
    \texttt{aida.kostikova@uni-bielefeld.de}
}

\begin{document}

\maketitle


\begin{abstract}


 
Solidarity is a crucial concept to understand social relations in societies. In this paper, we 
explore fine-grained solidarity frames to study solidarity 
towards women and migrants in German parliamentary debates between 1867 and 2022. Using 2,864 manually annotated text snippets (with a cost exceeding 18k Euro), we evaluate large language models (LLMs) like Llama 3, GPT-3.5, and GPT-4. We find that GPT-4 outperforms other LLMs, approaching human annotation quality. 
Using GPT-4, we automatically annotate more than 18k further instances (with a cost of around 500 Euro) across 155 years and find that solidarity with migrants outweighs anti-solidarity but that frequencies and solidarity types shift over time. Most importantly, group-based notions of (anti-)solidarity fade in favor of compassionate solidarity, focusing on the vulnerability of migrant groups, and exchange-based anti-solidarity, focusing on the lack of (economic) contribution. 
Our study highlights the interplay of historical events, socio-economic needs, and political ideologies in shaping migration discourse and social cohesion. 
We also show that powerful LLMs, if carefully prompted, can be cost-effective alternatives to human annotation for hard social scientific tasks.

\end{abstract}

\section{Introduction}\sectionlabel{sec:introduction}

\begin{table*}[!htb]
    \centering
    {\footnotesize
    \begin{tabularx}{\textwidth}{>{\hsize=0.6\hsize\raggedright\arraybackslash}X>{\hsize=1.4\hsize}X}
        \toprule
        \textbf{Gold Standard} & \textbf{Translation of the Original German Text} \\
        \midrule
        (1) \textbf{Compassionate solidarity towards women \newline{} (June 29, 1961)}
        & \enquote{In connection with § 1708 BGB, the Bundestag has set the age of 18 as the limit for the obligation to provide maintenance. In the transitional provisions, this stipulation has been repealed for those who had already reached the age of 16 on January 1, 1962. My faction finds this regulation unfair, as it would exempt significant groups of people from this maintenance obligation. \textbf{Especially women who have made great efforts to send their children to higher education, for example, would have to bear these costs alone.} [...]} \\
        \midrule
        (2) \textbf{Exchange-based \newline{} anti-solidarity towards migrants \newline{} (Apr. 19, 2018)}
        & \enquote{[...] \textbf{Let me also add: Migration is not necessarily successful -- you always act as if that is great -- it can fail, and it fails in particular when the immigrants' qualifications are low.} In 2013, before the so-called refugee wave, 40 percent of immigrants from non-EU countries had no qualifications. Since the wave of refugees, stabbings have increased by 20 percent, and we have imported anti-Semitism in the country. Does this make for an outstandingly successful migration?} \\
        \midrule
        (3) \textbf{Mixed stance towards migrants \newline{} (Feb. 2, 1982)}
        & \enquote{[...] We must accept that in a few years we will again need a higher number of foreign workers in the Federal Republic, as Mr. Urbaniak hinted earlier. \textbf{In reality, therefore, we must commit to effective integration, which admittedly requires [...] that there can be no exceptions, no alternative, regarding the recruitment stop and the prevention of illegal immigration}. [...]} \\
        \midrule
        (4) \textbf{None case (women) \newline{} (June 17, 2015)}
        & \enquote{[...] \enquote{We want to be free people!} There is probably no better phrase to open today's debate here in the German Bundestag about the popular uprising of 1953. [...] \textbf{We remember women and men who, 62 years ago, showed great courage because they wanted to change the course of their country's development and their own lives, because they wanted to be free people.}} \\
        \bottomrule
    \end{tabularx}}
    \caption{Example sentences from our dataset showing (anti-)solidarity towards women/migrants. Bold text is the main sentence, the other sentences are for context. Original German texts, as well as examples of mixed stance and none\se{,}  
    are available in Table~\ref{tab:solidarity-examples-de} in the Appendix.}
    \label{tab:solidarity-examples}
\end{table*}



\begin{figure}[!htb]
  \centering \includegraphics[width=0.9\linewidth]{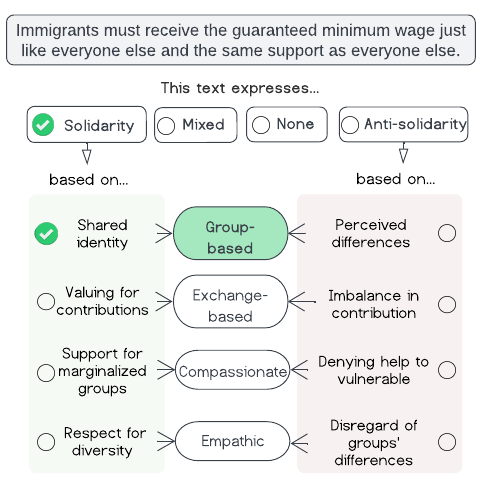}
  \caption{Annotation scheme based on \citet{thijssen2012mechanical}. The scheme categorizes statements into \emph{solidarity}, \emph{anti-solidarity}, \emph{mixed}, and \emph{none} (high-level). At the fine-grained level, \emph{solidarity} and \emph{anti-solidarity} are further divided into \emph{group-based}, \emph{exchange-based}, \emph{compassionate}, and \emph{empathic} subtypes.}
  \label{fig:annotation-scheme}
\end{figure}

Solidarity is a crucial concept for understanding how societies achieve and maintain  stability and cohesion \citep{reynolds2014introduction}, and it plays a critical role in shaping policies \citep{laitinen2014solidarity}. Traditionally, solidarity relied on common identity and reciprocity, potentially excluding out-groups like migrants \citep{hopman2022understanding} and reinforcing social boundaries and hierarchies \citep{anthias2014beyond}. However, growing diversity and increasing socio-economic, political, religious, and cultural complexities of modern societies \citep{kymlicka2020solidarity} call such traditional forms of solidarity into question. Simultaneously, recent populist movements challenge policies that support equality, such as equal opportunity or reproductive rights of women \citep{inglehart2016trump}.
These evolving complexities motivate a deeper and broader study of social solidarity, namely 
(i) a fine-grained exploration of different forms of social solidarity to reflect its multifaceted nature \citep{oosterlynck2013social}, and (ii) a broader historical analysis to trace its evolution from the 19th century to today \citep{banting2017strains}.
In this work, we contribute to such a systematic study of solidarity by tracing fine-grained notions of solidarity and anti-solidarity towards two 
target groups, women and migrants, in political speech, namely German parliamentary debates from 1867 to 2022 \citep{DeuParl}. 
We employ the social solidarity framework by \citet{thijssen2012mechanical} that incorporates rational (\emph{group-based} and \emph{exchange-based}) and emotive (\emph{compassionate} and \emph{empathic}) elements of solidarity
(refer to Fig.~\ref{fig:annotation-scheme} and \ref{sec:annotation} for more details on the typology). We focus on migrants, central for solidarity discourse in European and German politics \citep{thranhardt1993ursprunge, Faist1994, Froehlich2023, lehr2015germany}, and women as an \enquote{oppressed majority} historically marginalized from public life \citep{calloni2020women}. 
As manual annotation of (anti-)solidarity concepts in all parliamentary proceedings over this 155-year period using traditional sociological methods is practically infeasible, we explore the use of language models for this complex task. In particular, we assess the efficacy of BERT, Llama-3, GPT-3.5, and GPT-4, to detect expressions of various (anti-)solidarity types 
in parliamentary texts, aiming to identify the best performing model for our large-scale analysis. From an NLP perspective, this task is semantically and pragmatically challenging because, 
(i) expressions of (anti-)solidarity are often implied rather than explicitly stated in the text and their meaning is affected by the political and historical context in which they are made
\citep[see the examples in Table~\ref{tab:solidarity-examples};][]{sravanthi2024pub}; (ii) German data, especially evolving German language over 155 years, is under-represented in common training data sets for LLMs, which may affect performance  \citep{ahuja2023mega, qin2024multilingual, liu2024translation}; and
(iii) LLMs might struggle with annotating complex sociological concepts, achieving lower quality and reliability compared to human annotators \citep{wang2021want, ding2022gpt, zhu2023can, pangakis2023automated}.

Our contributions are: (i) We provide a human annotated training \& evaluation dataset of 2,864 text snippets,
which required 40+ hours weekly from 4-5 annotators over nine months, totaling an investment of approximately 18k Euro; 
(ii) we conduct a comparative analysis of LLMs on a complex sociological task 
in which pre-trained language models (esp. GPT-4) outperform an open-source model Llama-3-70B-Instruct, as well as models fine-tuned for this task (BERT, GPT-3.5 fine tuned); (iii) we provide fine-grained insights into solidarity discourse concerning migrants in Germany in the last 155 years across different political parties. 

\se{We make our code and data available at \url{https://github.com/DominikBeese/FairGer}.}
\section{Related work}\sectionlabel{sec:related_work}

Our work connects to (i) 
computational social science (CSS) 
(ii) analysis of political data (parliamentary debates) 
and (iii) the emergent field of analysis of social solidarity using 
NLP approaches. 

\paragraph{NLP-based CSS.}
Recent CSS studies have leveraged
LLMs for a variety of complex tasks. \citet{ziems2024can} conduct 
a comprehensive evaluation of LLMs, 
pointing out their weaknesses in tasks which require understanding of subjective expert taxonomies that deviate from the training data of LLMs (such as implicit hate and empathy classification). 
LLMs enhance text-as-data methods in social sciences, particularly in analyzing political ideology \citep{wu2023large}, but struggle with social language understanding, often outperformed by fine-tuned models \citep{choi2023llms}. \citet{zhang2023skipped} introduced SPARROW, a benchmark showing ChatGPT's limitations in sociopragmatic understanding across languages. 
In exploring German migration debates, \citet{blokker-etal-2020-swimming} and \citet{zaberer-etal-2023-political} utilize fine-tuning of transformer-based language models to classify claims in German newspapers. \citet{Chen2022-ew} apply LLM-based classification on German social media posts to study public controversies over the course of one decade.
In contrast to these approaches, we apply LLMs to longitudinal historical data and explore it for a new challenging task, fine-grained detection of social solidarity.

\paragraph{Analysis of parliamentary debates using NLP tools.}

\citet{abercrombie2020sentiment} review 61 studies on sentiment and position-taking within parliamentary contexts, covering 
dictionary-based sentiment scoring, statistical machine learning, and other conventional NLP methods. 
In terms of specific methodologies, studies often deploy: 
(i) shallow classifiers, where \citet{lai2020multilingual} use SVM, Naïve Bayes, and Logistic Regression for multilingual stance detection; (ii) deep learning approaches, with \citet{abercrombie2020parlvote} applying BERT, 
\citet{al2022sentence} exploring LSTM variants, and \citet{sawhney2020gpols} introducing GPolS for political speech analysis; (iii) probabilistic models, as in \citet{vilares2017detecting}'s Bayesian approach to identify topics and perspectives in debates. With German political debates, 
\citet{muller2021cares} use topic modeling to study shifts in German parliamentary discussions on coal due to changes in energy policy, while \citet{DeuParl} employ diachronic word embeddings to track antisemitic and anti-communist biases in these debates. More recently, \citet{bornheim2023speaker} 
apply Llama 2 to automate speaker attribution in German parliamentary debates from 2017-2021. 
Our research goes beyond this by adopting recent powerful LLMs to track changes of a specific social concept, solidarity, in plenary debates from three centuries.

\paragraph{Social solidarity in NLP.} 
Previous studies of social solidarity in NLP have largely focused on 
social media platforms. For example, \citet{Santhanam2019ISW} study how emojis are used to express solidarity in social media during Hurricane Irma in 2017 and Paris terrorist attacks from November 2015. \citet{TwitterDataset} consider solidarity in European social media discourse around COVID-19. 
\citet{Eger2022MeasuringSS} extend this work by examining how design choices, like keyword selection and language, affect assessments of solidarity changes over time. 
Compared to these works, we use a similar methodological setup (annotate data and infer trends), but 
focus on parliamentary debates instead of social media, employ a much more fine-grained sociological framework 
\citep{thijssen2012mechanical}, and use 
LLMs for systematic categorization and examination of solidarity types over time.

\section{Data}\sectionlabel{sec:data}

We obtain data from two sources: (i)~\textit{Open~Data}, covering \de{Bundestag}{federal diet} protocols from 1949 until today; and (ii)~\textit{Reichstagsprotokolle} covering \de{Reichstag}{imperial diet} protocols until 1945.
\footnote{\de{Volkskammer}{Eastern German parliament} protocols could not be included due to lack of availability.}
We use the OCR-scanned version from \citet{DeuParl}. Links to data, models, etc.\ used are in Appendix~\ref{sec:appendix-links}.
%
\subsectionlabel{sec:parliament-data}
For the \de{Reichstag} data, we apply preprocessing steps similar to \citet{DeuParl} (e.g., removal of OCR artifacts), but keep German umlauts, capitalization, and punctuation. We automatically split the data into individual sittings and collect metadata like the date, period and 
number of each sitting, which we manually check and correct. 
Additionally, we removed interjections and split the text into sentences using NLTK \citep{nltk}, resulting in 19.1M sentences. We release this dataset of plenary protocols from German political debates (DeuParl) consisting of 9,923 sittings from 1867 to 2022 on GitHub.
\footnote{\url{https://github.com/DominikBeese/DeuParl-v2}}



To select keywords, we (i) train a Word2Vec model \citep{Word2Vec} on our dataset to identify words with vector representations similar to \de{Migrant}{migrant} and \de{Frau}{woman}; (ii) manually expand this list with 
intuitively relevant terms; (iii) from both lists, we filter for those which appear at least 
200 times in the dataset. This resulted in 32 keywords for \de{Migrant} and 18 keywords for \de{Frau}. These include general terms like \de{Migrant}, \de{Immigrant} and \de{Frau} to period-specific 
terms, 
such as \de{Vertriebene}{expellees} and \de{Bürgerkriegsflüchtlinge}{civil war refugees}, or social roles, such as \de{Mütter}{mothers} and \de{Hausfrauen}{housewives}. 
See the full list of keywords, and further preprocessing in Appendix~\ref{sec:appendix-keywords}. 
For a detailed keyword distribution 
across the dataset, see Fig.~\ref{fig:frau-keywords-per-year} and Fig.~\ref{fig:migrant-keywords-per-year} in the Appendix.

Using these keywords, we extract 58k main sentences (\emph{instances}) for migrants and 131k for women from DeuParl, expanding each with three preceding and three following sentences for context, resulting in a total of 
(i) 463k sentences (9.79M tokens) for migrants and (ii) 1.58M sentences (32.82M tokens) for women. Fig.~\ref{fig:instances-per-year} shows the number of instances 
over time.\footnote{We note that the dataset is sparse in the period from 1933 to 1949, i.e.\ during the NS dictatorship and the immediate after-war period until the first parliament after the war was elected in 1949.} \ak{Fig.~\ref{fig:relative-frequency-records} in the Appendix shows yearly relative frequencies of sentences with terms related to women and migrants in the entire dataset. It is notable that both \de{Frau} and \de{Migrant} terms represent a minor fraction of the discourse, typically under 0.02. Periodic spikes in mentions likely align with historical and societal changes, such as post-WWII for \de{Migrant}.}


\vspace{-0.05cm}
\begin{figure}[!htb]
  \centering
  \includegraphics[width=0.96\linewidth]{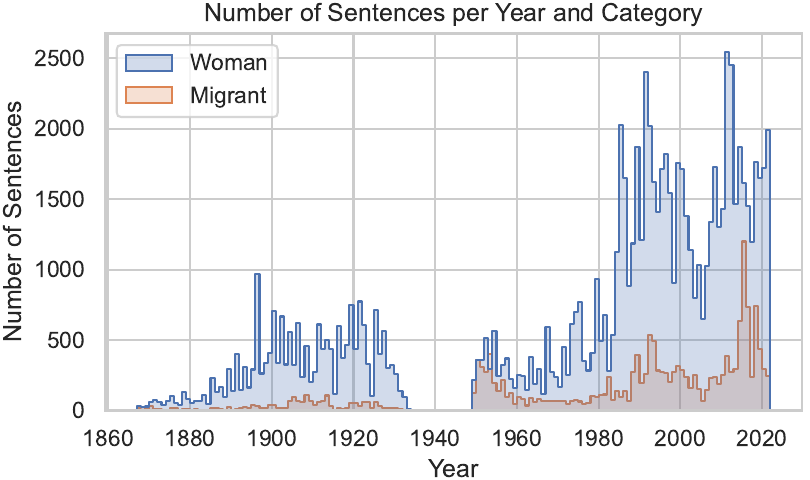}
  \caption{Number of instances in the \Frau{} and \Migrant{} dataset in each year. Fig.~\ref{fig:relative-frequency-records} in the Appendix illustrates the relative frequency of instances in both datasets.}
  \figurelabel{fig:instances-per-year}
\end{figure}
\vspace{-0.35cm}

\section{Data annotation}\sectionlabel{sec:annotation}

To obtain ground truth data for model training and evaluation, we annotated 2864 instances 
with five annotators (all student assistants, with specializations in social science or computer science). The annotation was performed over a duration of nine months. In the first three months, we iteratively refined the annotation guidelines and monitored the inter-rater agreement (measured by Cohen's Kappa) until inter-rater agreement converged 
(see \ref{sec:annotation-results} for exact scores) and annotators began annotating independently. 

\subsection{Annotation task design}\subsectionlabel{sec:annotation-task-design}

For the manual annotation, we take the target sentence and three preceding and following sentences for context into account. We first select a high-level category (\emph{solidarity}, \emph{anti-solidarity}, \emph{mixed}, \emph{none}). \emph{Solidarity} or \emph{anti-solidarity} cases are then further distinguished 
into frames as defined by \citet{thijssen2022s}: \emph{group-based}, \emph{compassionate}, \emph{exchange-based}, and \emph{empathic}. 
We describe each of the included variables below.

\paragraph{High-level categories.}


Based on \citet{Lahusen.2018} and \citet{TwitterDataset}, we define \emph{solidarity} as willingness to share resources, directly or indirectly, or support for target groups, and \emph{anti-solidarity} as statements restricting resources, showing unwillingness to support, or implying exclusion of these groups. Texts with both supporting and opposing expressions are labeled \emph{mixed}, while neutral or unrelated texts are labeled \emph{none}.

\paragraph{Group-based solidarity}
is coded for texts emphasizing shared identity and common goals among group members, whereas \emph{group-based anti-solidarity} emphasizes out-group exclusion 
based on perceived differences.

\paragraph{Compassionate solidarity}
is coded for texts supporting marginalized groups, emphasizing their need for protection, while \emph{compassionate anti-solidarity} dismisses these groups by considering them already in a good position, minimizing their need support or protection.

\paragraph{Exchange-based solidarity}
is coded when texts highlight the economic contributions of \enquote{exchange partners} and potential rewards or further contributions. Conversely, \emph{exchange-based anti-solidarity} calls for penalizing groups perceived as receiving more than they contribute. 

\paragraph{Empathic solidarity}
is coded when a speaker expresses respect for individual differences, seeing social diversity as beneficial, while \emph{empathic anti-solidarity} arises when differences are used as grounds for exclusion or neglect.


Annotation involved elaborate explanations, with identification of (anti-)solidarity resources and highlighting expressions of (anti-)solidarity, as well as \emph{self-} (speaker's own viewpoint) and \emph{other-position} (addressing or criticizing others' viewpoints). The full annotation process and a detailed example are described in Appendix~\ref{sec:annotation-details} and Fig.~\ref{fig:annotation-example} in the Appendix, respectively. 


\subsection{Annotation results}\subsectionlabel{sec:annotation-results}

\begin{figure}[!htb]
    \centering
    \subfloat[Annotators' agreement]{
      \includegraphics[width=0.51\linewidth]{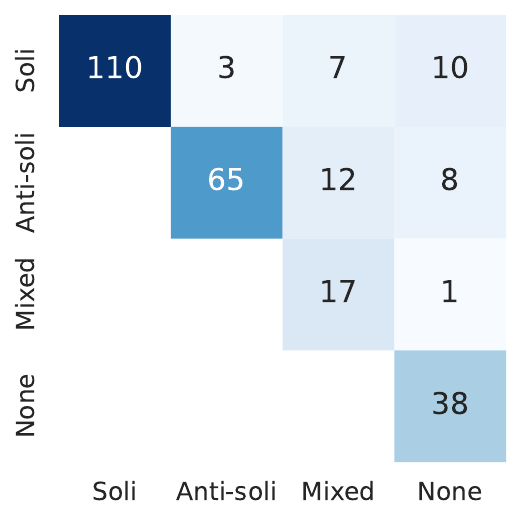}
        \label{subfig:annotation-confusion-matrix-highlevel}
    }
    \subfloat[Annotators vs. Model]{
        \includegraphics[width=0.475\linewidth]{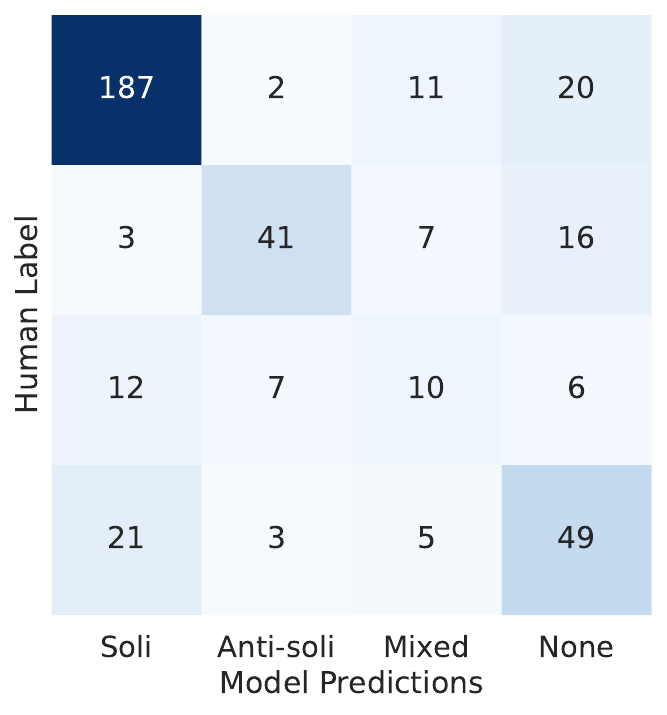}
        \label{subfig:confusion-matrix-highlevel}
    }
    \vspace{-0.1cm}
    \caption{Fig.~\ref{subfig:annotation-confusion-matrix-highlevel} \ak{shows \se{the} confusion matrix between human annotators; 
    Fig.~\ref{subfig:confusion-matrix-highlevel} shows agreement between the best model and human annotators on a test set from one of the three splits}. The former is aggregated over all pairwise comparisons of annotators, thus the matrix is symmetric.}
    \figurelabel{fig:CM-highlevel}
\end{figure}
\vspace{-0.05cm}
\todo{SE: I would rather say: Left shows confusion matrix between human annotators and right shows confusion matrix between best model and human. Btw. the labels are too small. Why is one symmetric and the other not? On which instances is this anyway? The y-axis is empty on the right side}
\todo{AK: fixed the labels; these are instances from a test set from one of the three splits; the axis on the right side is empty because these are the same labels in the same order as in the plot in the left}


While initial agreement levels were low, by the time annotators began working independently, they achieved a pairwise agreement with a Cohen's Kappa of 0.42 on a fine-grained level and 0.62 on a high level. 
We observe 
three main disagreement issues in annotation: 
misclassification of \emph{none} cases, confusion between \emph{mixed stance} and \emph{anti-solidarity}, and overlap within solidarity and anti-solidarity subtypes (see Fig.~\ref{fig:CM-finegrained} in the Appendix). This confusion is often due to overlapping characteristics or the presence of multiple subtypes within the text; \ak{moreover, this annotation task is inherently subjective, which can lead to differing interpretations. This is further evidenced by our average agreement scores.} Table~\ref{tab:annotator-divergence-examples} in the Appendix provides examples of annotator divergence, explaining why multiple labels could be correct, which gives insight into more difficult instances. However, 
there was almost no confusion 
between solidarity and anti-solidarity.
We note less stability in annotator agreement before 1930, stabilizing in subsequent years (see Fig.~\ref{subfig:kappa-over-time} in the Appendix). Although these variations can stem from the complexities of historical language and diverse interpretations of past events, they might also stem from the unbalanced distribution of human annotated data over the decades (see Fig.~\ref{fig:instances-distribution}).


Our dataset comprises 2864 annotated instances, 
1437 for migrants and 1427 for women. We note that anti-solidarity accounts for 13.5\% of instances, being more common among migrants (12.1\%) than women (1.4\%) (see Table~\ref{tab:label_distribution} in the Appendix). 
368 instances in our dataset (referred to as \emph{curated}) were reviewed by a social science expert to provide a reliable comparison benchmark for evaluation of our models. Other consensus mechanisms for the final labels in the human-annotated dataset, and their distribution are shown in Table~\ref{tab:levels_distribution} in the Appendix. 

\vspace{-0.1cm}
\begin{figure}[!htb]
  \centering
  \includegraphics[width=0.96\linewidth]{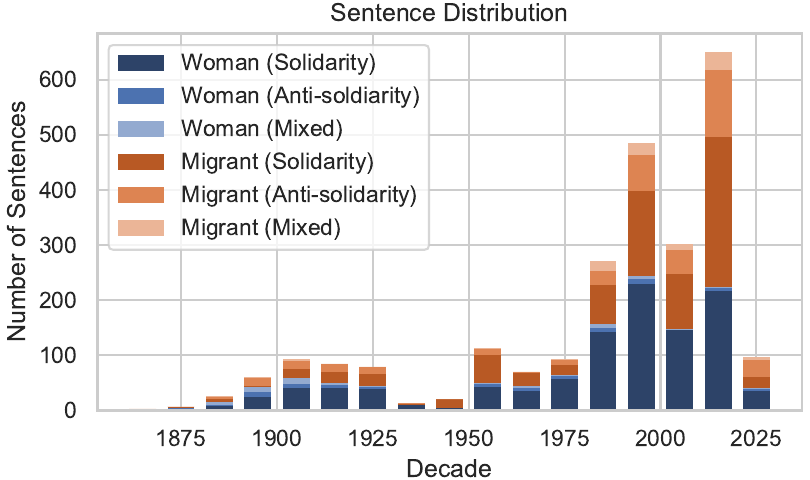}
  \caption{Distribution of instances in the human annotated dataset across time and target groups. See Fig.~\ref{fig:instances-distribution-categories} in the Appendix for the plots for each group separately; \ak{and Table~\ref{tab:actual-instances-distribution-categories} for the actual numbers of instances in the human annotated dataset.}}
  \figurelabel{fig:instances-distribution}
\end{figure}
\vspace{-0.1cm}
\section{Models and experiments}\sectionlabel{sec:models-experiments}


To determine the most cost-effective model (both in terms of performance and costs) for our large-scale sociological analysis, we evaluate various models, including \texttt{Llama-3-70B-Instruct}, \texttt{gpt-4-1106-preview}, base and instruction-finetuned \texttt{gpt-3.5-turbo-0125}, across high-level and fine-grained (anti-)solidarity annotation tasks in achieving human-level performance. 
Once the quality of the models is assured, we apply the best performing model --- GPT-4 --- large-scale to determine trends in \ref{sec:analysis}.


\paragraph{Data}

We use a 70/15/15 train/dev/test split for all \Migrant{} and \Frau{} annotated data, which gives us approx. 
2000 train, 400 dev and 430 test instances. 
We ensure reliability of a test set by allocating approximately 40\% curated and 45\% majority (\se{l}abels assigned when more than half of annotators agree on the same label) labels. 
We create 3 random 
data splits, and calculate performance metrics as the average score of the 3 runs on the test sets.\footnote{These sets are fully used for training and evaluating baseline models; for inference-based experiments with Llama-3, GPT-3.5, GPT-4, only test sets are used (also averaged on 3 runs).} 


\paragraph{Metrics}

To evaluate our models, %
we report the \textbf{Macro F1 Score (Macro F1)} to account for class imbalance. We also calculate the \textbf{F1 Score} for the classes individually. We report these 
metrics for both high-level and fine-grained tasks.


\subsection{Models}\subsectionlabel{sec:models}

\paragraph{Baseline}

For our baseline, we use a BERT-based pipeline with 110M parameters, comprising a high-level category classifier and two subtype models for \emph{solidarity} and \emph{anti-solidarity}. 
All models share a similar architecture, \todo{SE: Why models? Do you have multiple BERTs? Or does that hold for the below classifiers too? AK: yes, I train 3 different models, one just for solidarity, one just for anti-solidarity and one for high-level; this is more or less similar only to Llama-3. SE: well, but this is not stated anywhere?}
processing inputs with a target (\de{Frau} or \de{Migrant}) and full text \se{comprising the focus sentence and left and right context}. 
We add a fully connected layer with softmax activation atop the pooled output of the BERT-based models, with 4 output units 
for each model. To counter class imbalance, minority classes are oversampled to parity with the majority class. 
We finetune for 20 epochs with a learning rate of 4e-4, a warmup ratio of 0.05, linear decay, AdamW optimizer \citep{AdamW} and categorical crossentropy loss.


\paragraph{GPT-4}

We design two prompts (one for each target group) that include several elements: (i) incorporating chain-of-thought reasoning \citep{wei2022chain} in a few-shot setting by providing examples with desired reasoning and asking to \textit{think step by step} in a zero-shot setting \citep{kojima2022large}; 
(ii) providing precise definitions and insights derived from annotation discussions; 
(iii) introducing definitions and examples of potentially problematic labels (such as \emph{empathic solidarity} and \emph{empathic anti-solidarity}) earlier in the prompt 
and (iv) implementing a two-step prompting strategy that initially categorizes texts at a high-level followed by detailed subtype classification 
(full prompts are provided in Fig.~\ref{fig:prompt-migrant-appendix} and Fig.~\ref{fig:prompt-frau-appendix} in the Appendix).

\vspace{-0.1cm}

\paragraph{Prompt-based fine-tuned GPT-3.5}
%
Using the prompt identified for GPT-4's fine-grained classification, we proceeded to fine-tune GPT-3.5 on instances sampled from our initial train set (114 for migrants; 109 for women\footnote{The fine-tuning guide by OpenAI recommends using 50 to 100 examples for training: \url{https://platform.openai.com/docs/guides/fine-tuning/preparing-your-dataset}}), ensuring a balanced distribution across labels. The fine-tuning dataset was structured with the \textbf{system} providing instructions for classifying texts into high-level categories and requesting further sub-categorization; the \textbf{user} presenting texts; and the \textbf{assistant} providing classifications as per our two-step reasoning approach, along with explanations generated using GPT-4.

\paragraph{Llama-3}

For an open-source model, we select \texttt{Llama-3-70B-Instruct} with $q6\_k$ quantization \citep{meta2024introducing}, tested in a zero-shot 
setting. We use the same prompt as 
detailed in Fig.~\ref{fig:prompt-migrant-appendix} and Fig.~\ref{fig:prompt-frau-appendix} in the Appendix, but involving separate calls for high-level and subcategory classifications. 

For all GPT experiments, we test the models in few- (where we include ten category examples in the prompt to demonstrate the desired reasoning and categorization setting) and zero-shot settings. 


\begin{table*}[htb!]
    \centering
    \renewcommand{\arraystretch}{1.0}
    \setlength{\tabcolsep}{4pt}
    \footnotesize
    \begin{tabular}{c|cc|cc|c|c|c}
    \toprule
        \multirow{2}{*}{\shortstack{Model}} & \multicolumn{2}{c|}{\texttt{GPT-4}} & \multicolumn{2}{c|}{\shortstack{\texttt{GPT-3.5 base}}} & \multirow{2}{*}{\shortstack{\texttt{Llama-3-70B}\\0-shot}} & \multirow{2}{*}{\texttt{BERT}} & \multirow{2}{*}{\shortstack{Human\\upper bound}} \\
        & 0-shot & few-shot & 0-shot & few-shot & & & \\
    \midrule
        W & \textbf{0.37 (0.60)} & 0.37 (0.54) & 0.15 (0.46) & 0.12 (0.41) & 0.24 (0.48) & 0.13 (0.26) & 0.48 (0.72) \\
        M & \textbf{0.42 (0.73)} & 0.43 (0.63) & 0.19 (0.48) & 0.27 (0.50) & 0.27 (0.65) & 0.24 (0.46) & 0.56 (0.78) \\
    \bottomrule
    \end{tabular}
    \caption{Comparative performance (macro F1) of models vs. human upper bound on combined high-level (in parentheses) and fine-grained tasks for both women (W) and migrants (M). Best scores for each target group are highlighted in bold. Macro F1 cores for \texttt{GPT-3.5 fine-tuned}, as well as F1 scores for each category are provided in Table~\ref{tab:models_performance} in the Appendix.}
    \label{tab:models_performance_short}
\end{table*}

\section{Results}\sectionlabel{sec:results}

Results on the test sets are shown in Table~\ref{tab:models_performance_short} and Table~\ref{tab:models_performance} in the Appendix.

GPT-4 consistently outperforms other models in both fine-grained and high-level tasks for women and migrants. 
Interestingly, GPT-4 achieves similar performance in zero-shot and few-shot settings -- 0.37 (0.60) and 0.37 (0.54) for women, and 0.42 (0.73) and 0.43 (0.63) for migrants, respectively, where the first number represents the fine-grained, and the number in parentheses 
the high-level score. 
This might 
be attributed to the use of carefully crafted definitions, which eliminates the need for additional examples. 
The fine-tuned version of GPT-3.5 demonstrates only marginal improvement over the base model, generally falling short of GPT-4's performance. 
Most importantly, GPT-4 leads in F1 scores across categories, effectively identifying both solidarity and anti-solidarity (0.65 for women and 0.87 for migrants in zero-shot settings, see Table~\ref{tab:models_performance} in the Appendix). However, it scores lowest in the \emph{mixed stance} category, indicating challenges with more complex and ambiguous instances. Llama-3, on the other hand, outperforms GPT-4 in this category for women and also leads in the \emph{none} category for migrants by better identifying contradicting and subtle cues, likely benefiting from the two-step classification approach. 

Conversely, BERT achieves lower F1 scores across all categories, performing comparably to the GPT-3.5 versions on high-level tasks but struggling with fine-grained tasks and ambivalent categories like \emph{mixed stance}, likely due to challenges from underrepresented labels such as anti-solidarity categories for women, even after over-sampling (see Fig.~\ref{fig:instances-distribution}, as well as Table~\ref{tab:label_distribution} in the Appendix).

Overall, the human upper bound consistently outperforms all models. GPT-4 is the closest but still falls behind by 0.11 to 0.14 and 0.05 to 0.12 points in fine-grained and high-level tasks respectively for women and migrants. Llama-3-70B follows, with gaps of 0.24 to 0.29 and 0.24 to 0.13 points in fine-grained and high-level tasks, while BERT shows the largest discrepancies. These results suggest there is a gap compared to human understanding, especially in more complex annotation tasks.



\paragraph{Error analysis}\subsectionlabel{sec:error-analysis}

For the error analysis, we compare the human annotations and zero-shot predictions of GPT-4 for both target groups on the test set, using the confusion matrices for high-level labels shown in Fig.~\ref{fig:CM-highlevel}, as well as for fine-grained level labels in Fig.~\ref{fig:CM-finegrained} provided in the Appendix.
We also consider explanations provided by GPT-4.

Observed errors align with those noted during human annotation, with \emph{solidarity} and \emph{anti-solidarity} rarely confused (1\% of cases). Confusion primarily occurs between \emph{(anti-)solidarity} subtypes and \emph{none}, as well as  \emph{mixed stance}, as the model seems to look for stronger indications of solidarity 
despite instructions to consider even slight expressions of it (see examples 1 and 2 in Table~\ref{tab:predictions-examples}). 

There is also notable confusion between the solidarity subtypes, with the most frequent confusion 
between \emph{group-based} and \emph{compassionate solidarity},
likely because of the presence of multiple category characteristics within the texts (see example 3 in Table~\ref{tab:predictions-examples}). 
There is also confusion between \emph{compassionate} and \emph{empathic solidarity}, where the model sometimes misinterprets \emph{empathic} as merely relating to the term \emph{empathy}, overlooking the full category definition, which involves respecting diversity, as in example 4 in Table~\ref{tab:predictions-examples}. 

\begin{figure*}[!htb]
    \centering
    \subfloat[Solidarity and anti-solidarity frames.]{
        \includegraphics[width=0.57\linewidth]{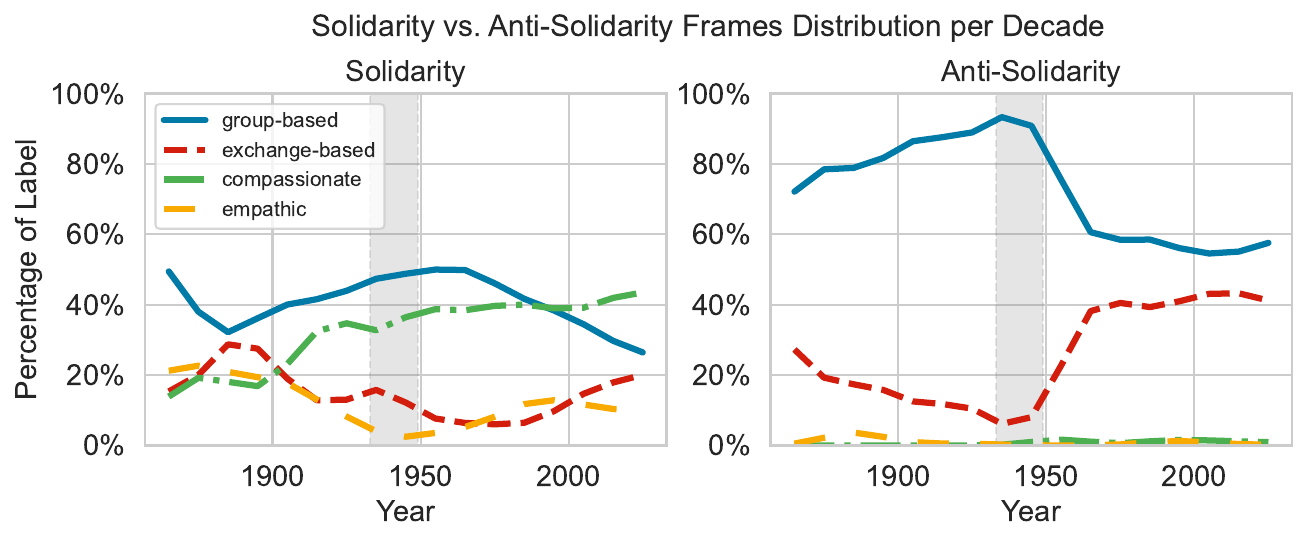}
        \label{fig:solidarity_antisolidarity}
    }
    \subfloat[Solidarity and anti-solidarity.]{
        \includegraphics[width=0.37\linewidth]{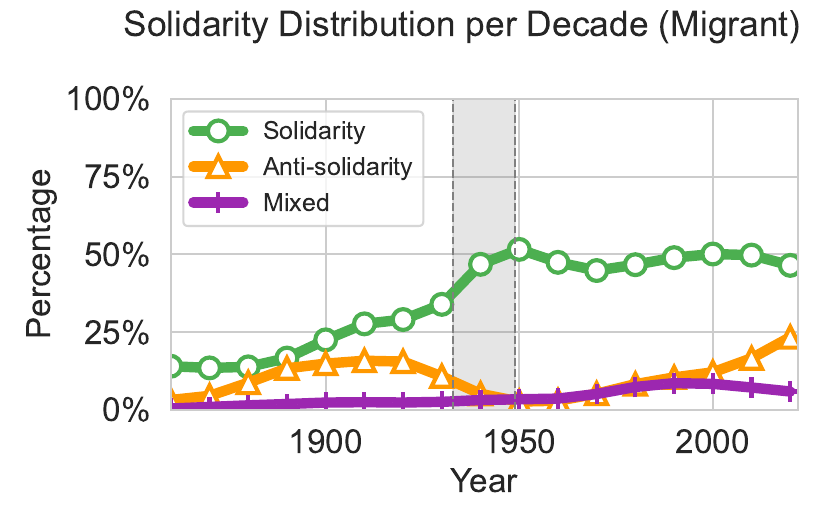}
        \label{fig:solidarity-per-decade}
    }
    \vspace{-0.1cm}
    \caption{Fig.~\ref{fig:solidarity-per-decade} shows the fraction of solidarity, anti-solidarity, and mixed stance towards migrants. Fig.~\ref{fig:solidarity-per-decade} shows the fraction of solidarity (left) and anti-solidarity (right) subtypes  according to GPT-4, where percentages represent the proportion of each subtype relative to the total counts of solidarity or anti-solidarity labels per decade.  Grey shaded areas from 1933 to 1949 indicate sparse data during the NS dictatorship and post-war period.}
\end{figure*}

\section{Analysis}\sectionlabel{sec:analysis}


In this section, we analyze how the solidarity discourse in the German parliament developed over 155 years, using automatic, fine-grained annotations from the best-performing model, GPT-4 zero-shot.
%
However, due to cost constraints, we limit the annotations to a) data concerning migrants, i.e., the target group for which our models achieved higher 
scores 
and b) a sample of 18,300 instances from the overall 58k instances concerning migrants.
We draw the sample proportionally for the time spans in the original data. 
This selection includes all records with political party information (see Appendix~\ref{sec:appendix-parties} for details on political parties data extraction and list of parties included in the analysis). \se{The cost of this automatic annotation was around 500 Euro.}


\paragraph{How does (anti-)solidarity change over time?}\subsectionlabel{sec:solidarity-trends}


As shown in Fig.~\ref{fig:solidarity-per-decade},  
throughout the periods analyzed, solidarity consistently surpasses anti-solidarity. From 1880 to 1910, solidarity increased from under 20\% to 30\%, driven by discussions about Eastern and Central European foreign workers' rights in the context of industrialization and local demographic shifts \citep{schonwalder1999invited}. 

Following the NS regime, solidarity surges above 50\%, aligning with the influx of \de{Vertriebene}{expellees}, who were generally viewed positively in parliamentary debates \citep{Froehlich2023}, as well as the arrival of \de{Gastarbeiter}{guest workers} from the Mediterranean \citep{stalker2000workers}. The highest solidarity and low anti-solidarity in 1940s reflect limited discussions (due to sparse data during this period) skewed towards solidarity, primarily focusing on expellees. 

Anti-solidarity towards migrants initially rose from about 5\% to over 15\% between 1870 and 1890, remaining stable until 1920, likely due to fears of \enquote{Polonization} from the influx of Polish workers, leading to 
restrictive policies against perceived threats to German identity \citep{triadafilopoulos2004building}. 
From 1960, anti-solidarity resurged with rising anti-migrant sentiments linked to labor migration in the 1960s and 1970s, right-wing opposition to liberal asylum laws in the 1990s \citep{Faist1994}, and the influx of refugees due to the Syrian war around 2015 with the subsequent rise of the extreme right-wing AfD party \citep{Hertner2022}. 
By 2022, solidarity stabilizes at around 40\%, while anti-solidarity has risen to above 20\%. We also note a relative decrease in solidarity compared to anti-solidarity, as shown in Fig.~\ref{fig:comparison-per-decade} in the Appendix, which illustrates this shift by subtracting the percentage of anti-solidarity from that of solidarity. 



\vspace{-0.1cm}

\paragraph{How have (anti-)solidarity frames evolved over time?}\subsectionlabel{sec:solidarity-trends-subtypes}

In Fig.~\ref{fig:solidarity_antisolidarity}, for solidarity (left), it is evident that group-based solidarity (i.e., emphasis on shared national identity and integration) was dominant until the 1990s, peaking to over 50\% in 1870 with the founding of the German empire. It drops to below 20\% by 1880, surges back above 50\% in 1970, and then declines to below 30\% by 2020. These trends align with periods of strong German nationalism (around 1870 and pre-World War II) and the influx of expellees in the 1950s and 1960s. 
For anti-solidarity (right), trends show group-based anti-solidarity dominant before WWII, peaking from 70\% to over 90\%, then declining to below 60\% post-war, reflecting a decline in opposition to migration based on national identity in parliamentary debates after the NS era. 
Instead, anti-solidarity arguments shifted to exchange-based anti-solidarity (from below 10\% to above 40\% after World War II), i.e.\ arguments that migrants are not providing adequate economic contributions. 
Neither compassionate nor empathic anti-solidarity are frequent at any time. 
By 2022, group-based solidarity declines, giving way to compassionate solidarity, which rose to above 40\%. 


\paragraph{How are solidarity and anti-solidarity frames represented across political parties?}\subsectionlabel{sec:party-trends}

Our analysis covers the distribution of (anti-)solidarity frames across parties from 1865 to 2022. It should be noted that we assess the data without regard to the specific emergence dates of political parties (e.g., AfD established in 2013 versus SPD in 1863).

Data subdivision by the speaker's party shows that all parties, except for the far-right AfD, 
mainly exhibit solidarity towards migrants (Fig.~\ref{fig:party-subtypes} in the Appendix, left). Compassionate and group-based solidarity are prevalent, with left-leaning parties (Linke, Grüne) showing higher levels of compassionate and empathic solidarity compared to the more exchange-focused solidarity of centrist parties (SPD, FDP, CDU/CSU). 
Conversely, while the right-wing parties, including CDU/CSU, also engage in anti-solidarity rhetoric, it is AfD that predominantly holds such stances, 
focusing mainly on exchange-based anti-solidarity that suggests migrants contribute less (see Fig.~\ref{fig:party-subtypes} in the Appendix, right). 
This distribution mirrors findings in Flanders, Belgium, where \ak{radical rightists\se{'} discourse predominantly focus\se{es} on group-based frames}; greens and social democrats emphasize compassionate solidarity, liberals prefer exchange-based approaches, and both greens and liberals advocate for empathic solidarity, highlighting a polarization of partisan discourse \citep{thijssen2022s}. 

\section{Concluding remarks}\sectionlabel{sec:conclusion}

This study set out to i) provide a high-quality dataset of (anti-)solidarity annotations in German parliamentary debates, ii) from an NLP perspective, evaluate the ability of LLMs to assist in large-scale social and political analyses over extended periods, and iii) from a CSS perspective, uncover long-term shifts in solidarity trends in Germany, laying groundwork for future sociological studies.

Concerning i), we invested 1000+ annotation hours and 18k Euro to provide a data set of 2,864 manually annotated text snippets 
following the sociological framework of \citet{thijssen2012mechanical}.

Regarding ii), our findings indicate that GPT-4 outperforms other models (Llama-3-70B, GPT-3.5 base, fine-tuned, and BERT) in reproducing human annotations. While other models demonstrate proficiency in identifying high-level categories, they exhibit limitations in handling more nuanced categories, specific to sociological theory. GPT-4, though challenged by ambiguity, handles complex tasks even in a zero-shot setting. 

Contrary to previous research in CSS that suggested smaller fine-tuned models like BERT performed well \citep{choi2023llms, zhang2023skipped}, our study 
\se{finds}
that larger models are more effective 
\se{our hard CSS task,} 
possibly due to the carefully designed prompts based on human expertise. 
Our observations align with \citet{ziems2024can}, who noted that while LLMs have not yet reached the quality of human analysis in classification tasks within CSS, large instruction-tuned LLMs are preferable. However, we observe that the open-source model Llama-3 exhibits inferior performance compared to the non-open-source GPT-4.

In terms of iii), our analysis of German parliamentary debates from 1865 to 2022 reveals shifts in attitudes toward migrants influenced by labor demands, migration waves, and socio-political changes. The evolution from group-based to compassionate solidarity marks a shift towards economic pragmatism and issues of redistribution in migration discourse, likely spurred by rising global humanitarianism \citep{vollmer2018volatility}. Despite traditional solidarity dominance, the resurgence of anti-solidarity since WWII, peaking in 2022, illustrates deepening political polarization. This underscores the ongoing tension between xenophobic tendencies and liberal ideals in Germany \citep{joppke2007beyond, lehr2015germany}, reflecting the complex challenges of integration and multiculturalism in a major migrant destination.

These findings lay a strong foundation for further sociological research using LLMs. Future work could examine how shifts in political discourse impact policy-making and public opinion, exploring correlations with civic solidarity. Comparative studies across countries with varying migration histories may reveal factors influencing (anti-)solidarity. Additionally, exploring the effects of such rhetoric on related policies, like migration or equal opportunity, could yield further insights. 
From an NLP perspective, future work can explore a broader range of LLMs, including those fine-tuned on German data, and explore the effect of using highlighting and explanations to improve classification. 




\section*{Limitations}\sectionlabel{sec:limitations}

Our study faces several limitations that should be considered when interpreting the results. 

\begin{itemize}
    \item The task of annotating political speech, particularly concepts such as solidarity and anti-solidarity, poses significant challenges. These concepts are inherently complex and laden with subtleties that are difficult to capture, both for human annotators and automated models. 
    \item Due to resource constraints, GPT-4 was applied to only a portion of the dataset, potentially limiting the generalizability of our findings. Moreover, the proportional sampling of instances across decades might have led to the underrepresentation of certain periods, which could have further impacted the comprehensiveness of the analysis.
    \item While our analysis of solidarity and anti-solidarity frames distribution across parties covers data from 1865 to 2022 and includes all historical periods, it does not trace the evolution of partisan discourse over time. We assess the data comprehensively without focusing on the specific historical emergence of political parties (e.g., AfD established in 2013 versus SPD in 1863).
    \item Additional manual annotation features such as free-comment explanations, highlighting, and indications of opposing positions \se{(all available in our human annotation)} were not fully explored. These elements hold potential for future studies, which could use them as cues for LLMs to achieve a more nuanced classification.
\end{itemize}



\section*{Acknowledgments}\sectionlabel{sec:acknowledgments}

This research was funded by the Ministry of Culture and Science of the State of North Rhine-Westphalia under the grant no NW21-059A (SAIL).
The NLLG group gratefully acknowledges support from the Federal Ministry of Education and Research (BMBF) via the research grant ``Metrics4NLG'' and the German Research Foundation (DFG) via the Heisenberg Grant EG 375/5-1.

\bibliography{bibliography}

\appendix
\appendix

\section*{Appendix}\label{sec:appendix}

\section{List of Keywords}
\label{sec:appendix-keywords}

For \de{Frau}{woman} we use Frauen, Frau, Mutter, Mädchen, Mütter, Ehefrau, Müttern, Hausfrauen, Hausfrau, Ehefrauen, Frauenförderung, Frauenquote, Dienstmädchen, Fräulein, Großmutter, Kriegerfrauen, Arbeiterfrauen, and Trümmerfrauen.

Since \de{Frau} is the German word for \textit{woman} but also for \textit{Mrs./Ms.}, we only include occurrences of the word \de{Frau} that are not followed by a capitalized word (which are probably surnames).

For \de{Migrant}{refugee} we use Flüchtlinge, Ausländer, Flüchtlingen, Zuwanderung, Vertriebenen, Ausländern, Asylbewerber, Migranten, Migration, Heimatvertriebenen, Aussiedler, Einwanderung, Ansiedler, Vertriebene, Zuwanderer, Asylbewerbern, Flüchtling, Heimatvertriebene, Sowjetzonenflüchtlinge, Aussiedlern, Einwanderer, Asylsuchenden, Asylsuchende, Bürgerkriegsflüchtlinge, Zuwanderern, Ansiedlern, Migrantinnen, Vertriebener, Emigranten, Kriegsflüchtlinge, Ausländerinnen, and Immigranten.

When doing stability tests over the chosen keywords, we make sure to choose sufficiently many keywords, i.e., at least 5 (out of 16/32) keywords and at least 10\% of the data, such that enough data is present to create the plots.
For the analysis of frequency of keywords over time, we calculate the percentages normalized for each keyword, i.e., a value of $p$\% in year $y$ implies that in year $y$ $p$\% of all sentences with this keyword occurred. The trends are shown in the Fig.~\ref{fig:frau-keywords-per-year} for women and in Fig.~\ref{fig:migrant-keywords-per-year} for migrants.


\section{Parties}
\label{sec:appendix-parties}

We identified political parties by searching for party names within parentheses, a conventional notation within parliamentary records to denote the speaker's party affiliation, as seen in examples like \textit{"Benjamin Strasser (FDP): Sehr geehrter Präsident!..."}. 
Following automated extraction, we conducted a manual review to verify the correctness of the party associations, which resulted in 3,499 out of 58k records with party information spanning from 1940 to 2022. 

List of the parties included in the dataset, along with the variations of their names or abbreviations as they have been recorded:
AfD (Alternative for Germany); Die Linke (The Left) with variations such as PDS, Gruppe der PDS; Bündnis 90/Die Grünen (Alliance 90/The Greens); CDU/CSU (Christian Democratic Union/Christian Social Union); SPD (Social Democratic Party of Germany); FDP (Free Democratic Party); DP (German Party) with variations such as DP/DPB, DP/FVP, FVP; GB/BHE (All-German Bloc/League of Expellees and Deprived of Rights); KPD (Communist Party of Germany); BP (Bavarian Party); WAV (Economic Reconstruction Union).

\section{Annotation details}
\label{sec:annotation-details}

Annotators first identify the specific resources (such as time, money, rights, or educational access) that underlie expressions of \emph{solidarity} or \emph{anti-solidarity}. Following this, they select from a predefined set of indicators tailored to each \emph{solidarity} or \emph{anti-solidarity} category. 

\begin{figure*}[!htb]
    \centering
    \begin{subfigure}[b]{1\textwidth}
        \centering
        \includegraphics[width=\linewidth]{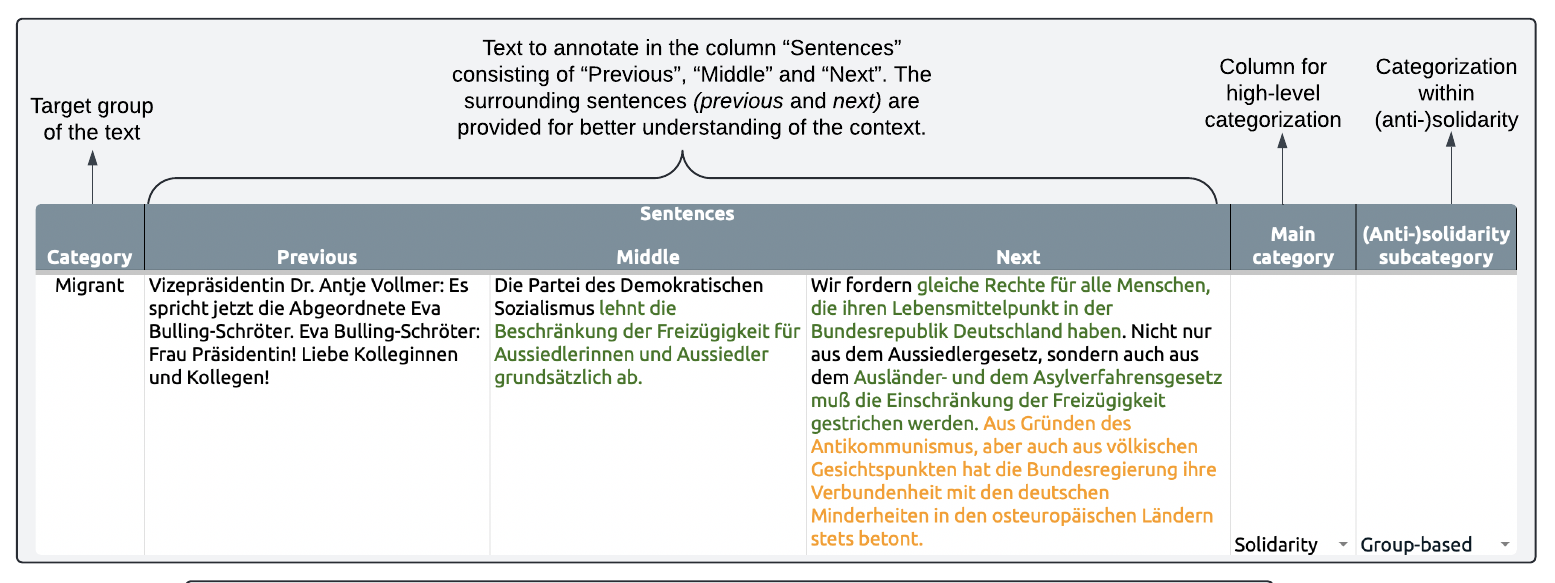}
        \caption{Columns for high-level and (anti-)solidarity categorizations.}
        \label{fig:annotation-example-1}
    \end{subfigure}
    
    \begin{subfigure}[b]{0.7\textwidth}
        \centering
        \includegraphics[width=\linewidth]{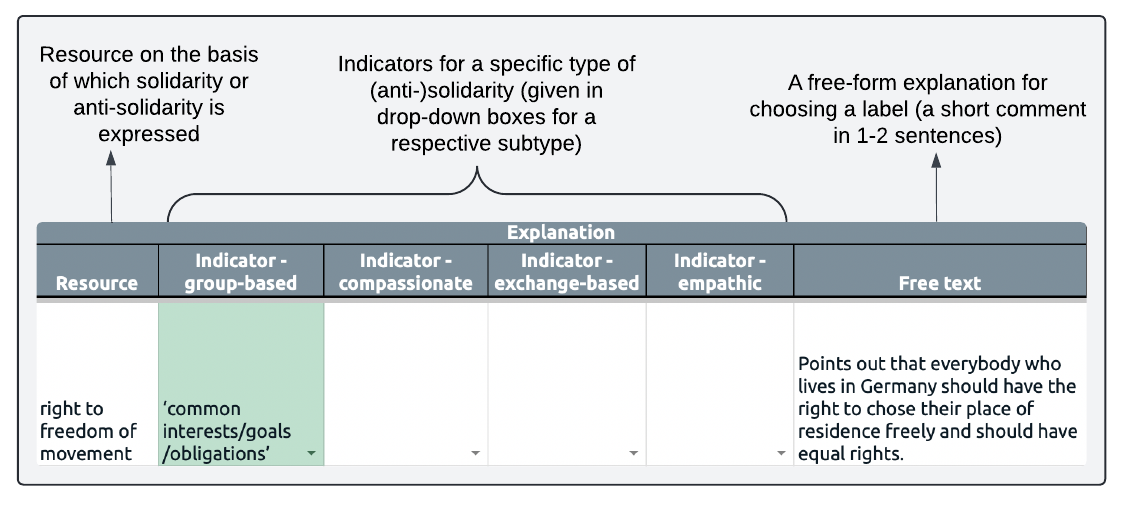}
        \caption{Columns for providing explanations.}
        \label{fig:annotation-example-2}
    \end{subfigure}
    
    \caption{Example of the annotation process from the annotation file. Fig.~\ref{fig:annotation-example-1} illustrates the step where annotators choose a high-level label and an (anti-)solidarity subcategory, if applicable. Fig.~\ref{fig:annotation-example-2} shows columns for detailed explanations, including the choice of a resource, an indicator, and providing free-text commentary.}
    \label{fig:annotation-example}
\end{figure*}

We also apply highlighting to indicate parts of sentences which express solidarity (green) and anti-solidarity red. In addition, we highlight expressions that convey the speaker's own viewpoint (\emph{self-position}) and those that address or critique others' viewpoints (\emph{other-position}). The annotation task concludes with a free-text commentary, limited to 1-2 sentences, detailing the reasoning behind a chosen category. 
A detailed example for the annotation process is available in Fig.~\ref{fig:annotation-example}, 
which illustrates the full annotation pipeline with providing explanations for chosen labels.


\section{Links to data and code}
\label{sec:appendix-links}

\textit{Open~Data}: \url{https://www.bundestag.de/services/opendata};
\textit{Reichstagsprotokolle}: \url{https://www.reichstagsprotokolle.de/};
OCR-scanned version of \citet{DeuParl}: \url{https://tudatalib.ulb.tu-darmstadt.de/handle/tudatalib/2889};
F1 score implementation: \url{https://scikit-learn.org/stable/modules/generated/sklearn.metrics.f1_score.html};

\begin{table*}[!htb]
    \centering
    {\footnotesize
    \begin{tabularx}{\textwidth}{>{\hsize=.6\hsize\raggedright\arraybackslash}X>{\hsize=1.7\hsize}X>{\hsize=.7\hsize}X}
        \toprule
        \textbf{Gold Standard} & \textbf{Original Text} & \textbf{Explanation} \\
        \midrule
        (1) \textbf{Compassionate solidarity towards women \newline{} (June 29, 1961)}
        & \enquote{Im Zusammenhang mit § 1708 BGB hat das Hohe Haus das 18. Lebensjahr als Grenze für die Unterhaltspflicht festgelegt. In den Übergangsvorschriften ist diese Bestimmung für diejenigen, die am 1. Januar 1962 schon das 16. Lebensjahr vollendet haben, aufgehoben worden. Diese Regelung erscheint meiner Fraktion ungerecht, denn dadurch würden beträchtliche Personengruppen aus dieser Unterhaltspflicht herausgenommen. \textbf{Gerade die Frauen, die unter großen Mühen ihre Kinder z. B. auf die höhere Schule geschickt haben, müßten diese Unkosten ganz allein tragen.}} & The speaker is advocating for extended financial support for mothers 
        and is emphasizing the unfairness of removing maintenance obligations. \\
        \midrule
        (2) \textbf{Exchange-based anti-solidarity towards migrants \newline{} (Apr. 19, 2018)}
        & \enquote{[...] \textbf{Lassen Sie mich noch anfügen: Migration ist nicht zwingend erfolgreich -- Sie tun immer so, als sei das super --, sie kann scheitern, und sie scheitert vor allem dann, wenn die Qualifikation der Einwanderer niedrig ist.} 2013, also vor der sogenannten Flüchtlingswelle, hatten 40 Prozent der Zuwanderer aus dem Nicht-EU-Ausland keinen Abschluss. Seit der Flüchtlingswelle haben die Messerstechereien um 20 Prozent zugenommen, und wir haben importierten Antisemitismus im Land. Ist das eine hervorragend erfolgreiche Migration?} & The text criticizes migration for its negative economic impacts and the disproportionate burden placed by low-skilled immigrants who take more resources and social stability than they contribute.\\
        \midrule
        (3) \textbf{Mixed stance towards migrants \newline{} (Feb. 2, 1982)}
        & \enquote{[...] Wir müssen akzeptieren, daß wir in wenigen Jahren auch wieder eine höhere Zahl ausländischer Arbeitnehmer in der Bundesrepublik brauchen werden, wie Herr Urbaniak vorhin angedeutet hat. \textbf{Wir haben also in Wirklichkeit zu einer wirksamen Integration, die allerdings voraussetzt, [...] daß es in der Frage des Anwerbestopps und des Verhinderns der illegalen Einwanderung keine Ausnahmen geben darf, keine Alternative.}. [...]} & This text acknowledges the economic need for foreign workers and the importance of their integration, yet simultaneously emphasizing strict controls on illegal immigration. \\
        \midrule
        (2) \textbf{None case (women) \newline{} (June 17, 2015)}
        & \enquote{[...] \enquote{Wir wollen freie Menschen sein!} Es gibt wohl keinen besseren Satz, um die heutige Debatte hier im Deutschen Bundestag über den Volksaufstand von 1953 zu eröffnen. [...] \textbf{Wir erinnern an Frauen und Männer, die vor 62 Jahren viel Mut bewiesen, weil sie der Entwicklung ihres Landes und ihrem eigenen Leben eine andere Richtung geben wollten, weil sie freie Menschen sein wollten.}} & The mention of women is integrated into the broader remembrance of the collective effort of people fighting against oppression without emphasizing any specific women's issues or needs.\\
        \bottomrule
    \end{tabularx}}
    \caption{Original German texts for the examples from our dataset in Table~\ref{tab:solidarity-examples} showing solidarity/anti-solidarity towards women/migrants. Bold text is the main sentence, the other sentences are for context.}
    \tablelabel{tab:solidarity-examples-de}
\end{table*}

\begin{figure*}[!htb]
    \centering
    \subfloat[Annotators' agreement]{
        \includegraphics[width=0.465\linewidth]{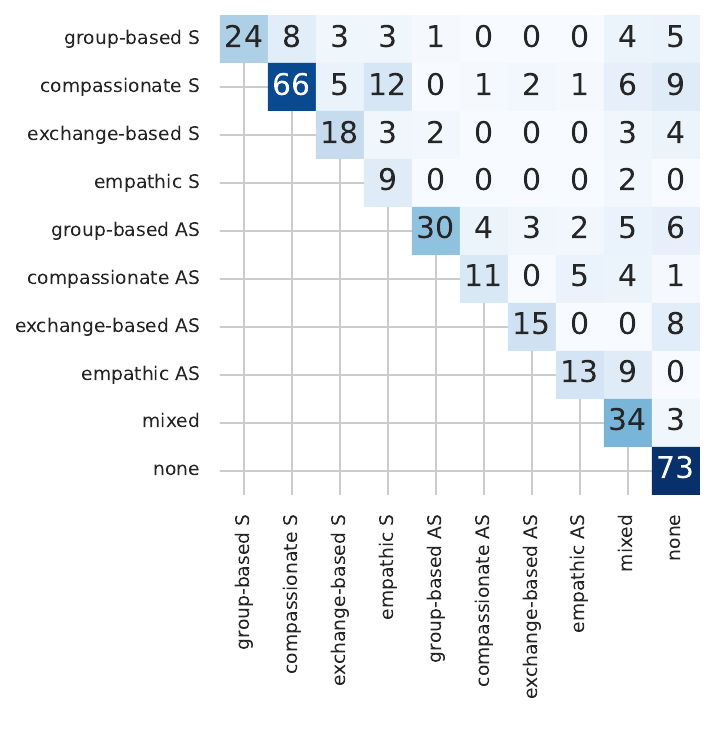}
        \label{subfig:annotation-confusion-matrix-finegrained}
}
    \subfloat[Annotators vs. Model]{
        \includegraphics[width=0.49\linewidth]{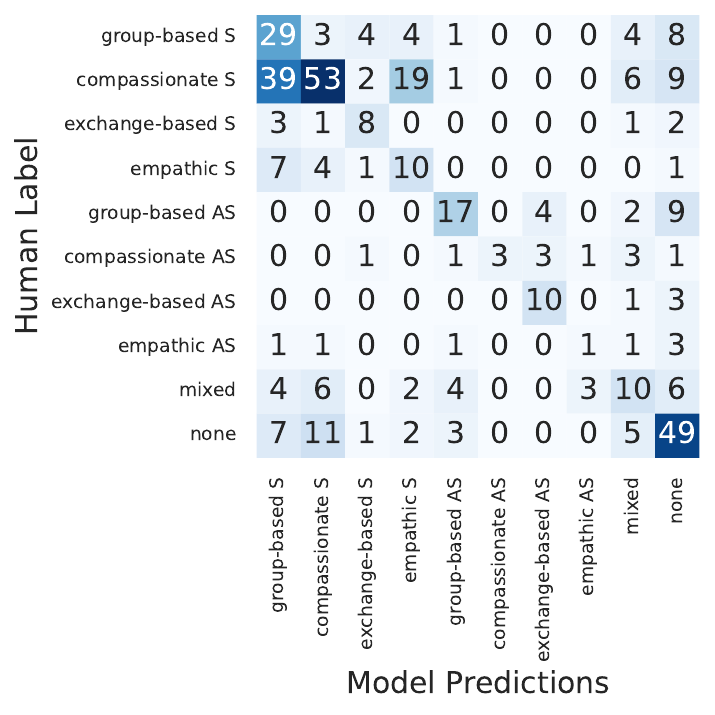}
        \label{subfig:confusion-matrix-finegrained}
    }
    \vspace{-0.1cm}
    \caption{Table~\ref{subfig:annotation-confusion-matrix-finegrained} shows the comparison of annotations between our annotators on a fine-grained level; Table~\ref{subfig:confusion-matrix-finegrained} between the final label from the human annotated dataset and our best model's prediction (cf.\ \ref{sec:results}) on a test set. The former is aggregated over all pairwise comparisons of annotators, thus the matrix is symmetric.}
    \figurelabel{fig:CM-finegrained}
\end{figure*}

\begin{figure*}[!htb]
    \centering
    \includegraphics[width=0.9\linewidth]{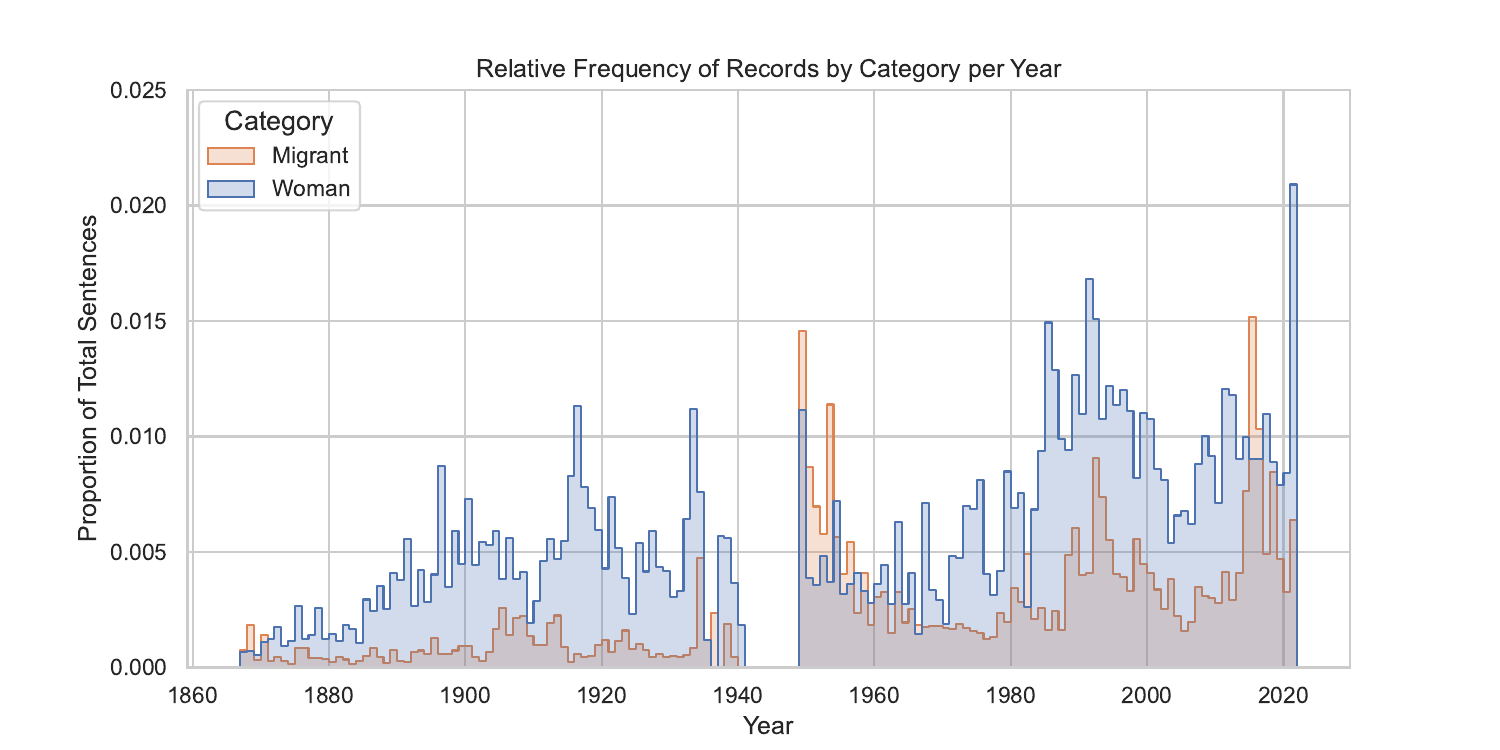}
    \caption{Relative frequency of instances per year in the \Frau{} and \Migrant{} dataset.}
    \figurelabel{fig:relative-frequency-records}
\end{figure*}

\begin{table*}[htb!]
    \centering 
    \small 
    \setlength{\tabcolsep}{4pt} 
    \begin{tabular}{@{}ccccc|cccc@{}}
        \toprule
        & \multicolumn{4}{c|}{Women} & \multicolumn{4}{c}{Migrants} \\
        \cmidrule(lr){2-5} \cmidrule(lr){6-9}
        Decade & S & AS & Mixed & None & S & AS & Mixed & None \\
        \midrule
        1860s & 1 & 1 & 0 & 10 & 7 & 1 & 0 & 9 \\
        1870s & 5 & 5 & 0 & 7 & 7 & 2 & 0 & 8 \\
        1880s & 8 & 2 & 6 & 6 & 6 & 5 & 3 & 6 \\
        1890s & 25 & 8 & 9 & 5 & 15 & 1 & 1 & 11 \\
        1900s & 41 & 6 & 12 & 21 & 16 & 14 & 4 & 11 \\
        1910s & 40 & 6 & 4 & 19 & 20 & 15 & 0 & 6 \\
        1920s & 38 & 4 & 2 & 16 & 22 & 13 & 0 & 7 \\
        1930s & 15 & 0 & 1 & 3 & 5 & 3 & 0 & 12 \\
        1940s & 12 & 0 & 1 & 6 & 8 & 0 & 0 & 8 \\
        1950s & 42 & 6 & 2 & 14 & 51 & 10 & 3 & 16 \\
        1960s & 36 & 4 & 5 & 19 & 24 & 1 & 0 & 6 \\
        1970s & 57 & 6 & 2 & 17 & 18 & 9 & 2 & 12 \\
        1980s & 142 & 7 & 8 & 26 & 71 & 24 & 19 & 17 \\
        1990s & 229 & 10 & 5 & 41 & 154 & 65 & 22 & 20 \\
        2000s & 146 & 1 & 1 & 27 & 100 & 43 & 11 & 9 \\
        2010s & 216 & 6 & 2 & 49 & 272 & 121 & 34 & 42 \\
        2020s & 35 & 5 & 1 & 6 & 20 & 30 & 6 & 11 \\
        \bottomrule
    \end{tabular}
    \caption{Actual numbers of instances in the human annotated dataset across time per target group.}
    \label{tab:actual-instances-distribution-categories}
\end{table*}

\begin{figure*}[!htb]
    \centering
    \includegraphics[width=\linewidth]{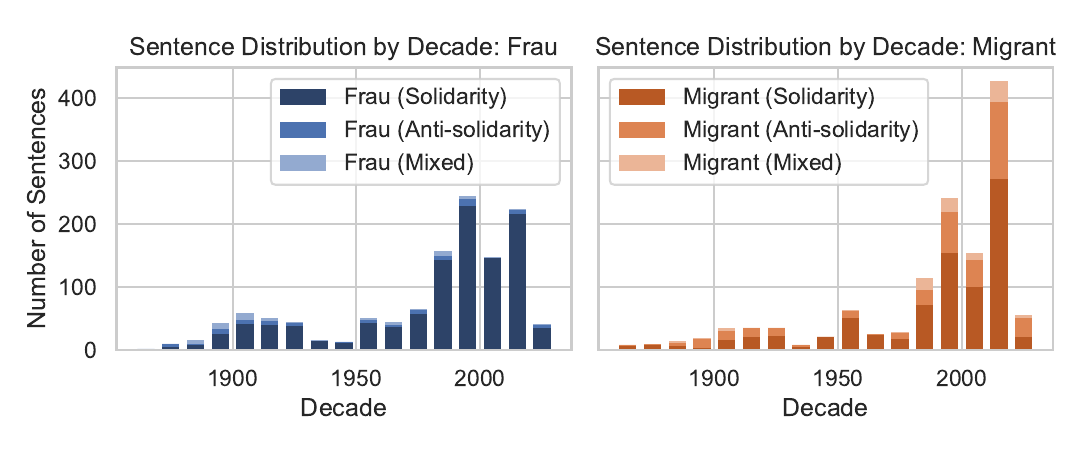}
    \caption{Distribution of instances in the human annotated dataset across time per target group.}
    \figurelabel{fig:instances-distribution-categories}
\end{figure*}

\begin{table*}[htb!]
\centering
    \begin{subtable}{\textwidth}
        \centering
        \begin{tabular}{l l r r r}
            \toprule
            & \textbf{Women} & \textbf{Migrant} & \textbf{Total per label} \\
            \midrule
            Group-based solidarity & 112 (3.9\%) & 188 (6.6\%) & 300 (10.5\%) \\
            Exchange-based solidarity & 54 (1.9\%) & 56 (2\%) & 110 (3.8\%) \\
            Empathic solidarity & 125 (4.4\%) & 21 (0.7\%) & 146 (5.1\%) \\
            Compassionate solidarity & 732 (25.6\%) & 466 (16.3\%) & 1198 (41.8\%) \\
            Solidarity (no subtype) & 41 (1.4\%) & 53 (1.9\%) & 94 (3.3\%) \\
            \addlinespace
            \textbf{Total for solidarity} & \textbf{1064 (37.2\%)} & \textbf{784 (27.4\%)} & \textbf{1848 (64.5\%)} \\
            \midrule
            Group-based anti-solidarity & 10 (0.3\%) & 197 (6.9\%) & 207 (7.2\%) \\
            Exchange-based anti-solidarity & 0 (0\%) & 48 (1.7\%) & 48 (1.7\%) \\
            Empathic anti-solidarity & 17 (0.6\%) & 3 (0.1\%) & 20 (0.7\%) \\
            Compassionate anti-solidarity & 8 (0.3\%) & 80 (2.8\%) & 88 (3.1\%) \\
            Anti-solidarity (no subtype) & 5 (0.2\%) & 19 (0.7\%) & 24 (0.8\%) \\
            \addlinespace
            \textbf{Total for anti-solidarity} & \textbf{40 (1.4\%)} & \textbf{347 (12.1\%)} & \textbf{387 (13.5\%)} \\
            \midrule
            \textbf{Mixed} & 60 (2.1\%) & 101 (3.5\%) & 161 (5.6\%) \\
            \textbf{None} & 273 (9.5\%) & 195 (6.8\%) & 468 (16.3\%) \\
            \addlinespace
            \textbf{Instances in total} & \textbf{1437 (50.2\%)} & \textbf{1427 (49.8\%)} & \textbf{2864} \\
            \bottomrule
        \end{tabular}
        \caption{Distribution of labels by target group.} 
        \label{tab:label_distribution}
    \end{subtable}
    \vspace{1em}

    \begin{subtable}{\textwidth}
    \centering
        \begin{tabular}{l r r r}
            \toprule
            \textbf{Label} & \textbf{Curated} & \textbf{Majority} & \textbf{Single} \\
            \midrule
            Group-based solidarity & 57 & 45 & 198 \\
            Exchange-based solidarity & 19 & 22 & 69 \\
            Empathic solidarity & 28 & 14 & 104 \\
            Compassionate solidarity & 119 & 202 & 877 \\
            Solidarity (no subtype) & 5 & 8 & 81 \\
            \addlinespace
            \textbf{Total for solidarity} & \textbf{228} & \textbf{291} & \textbf{1329} \\
            \midrule
            Group-based anti-solidarity & 20 & 53 & 134 \\
            Exchange-based anti-solidarity & 11 & 15 & 22 \\
            Empathic anti-solidarity & 1 & 18 & 1 \\
            Compassionate anti-solidarity & 1 & 29 & 58 \\
            Anti-solidarity (no subtype) & 2 & 3 & 19 \\
            \addlinespace
            \textbf{Total for anti-solidarity} & \textbf{35} & \textbf{118} & \textbf{234} \\
            \midrule
            \textbf{Mixed} & 21 & 40 & 100 \\
            \textbf{None} & 84 & 98 & 286 \\
            \addlinespace
            \textbf{Instances per label level (out of 2864)} & \textbf{368} & \textbf{547} & \textbf{1949} \\
            \bottomrule
        \end{tabular}
        \caption{Distribution of instances per label level. \emph{Curated}: labels established by manual revision by an expert; \emph{majority}: labels assigned when more than half of annotators agree on the same label; \emph{single}: instances with only one annotation.}
        \label{tab:levels_distribution}
        \end{subtable}
        
    \caption{Human annotated dataset statistics.}
    \label{tab:dataset-statistics}

\end{table*}

\begin{figure*}[!htb]
    \centering
    \includegraphics[width=\linewidth]{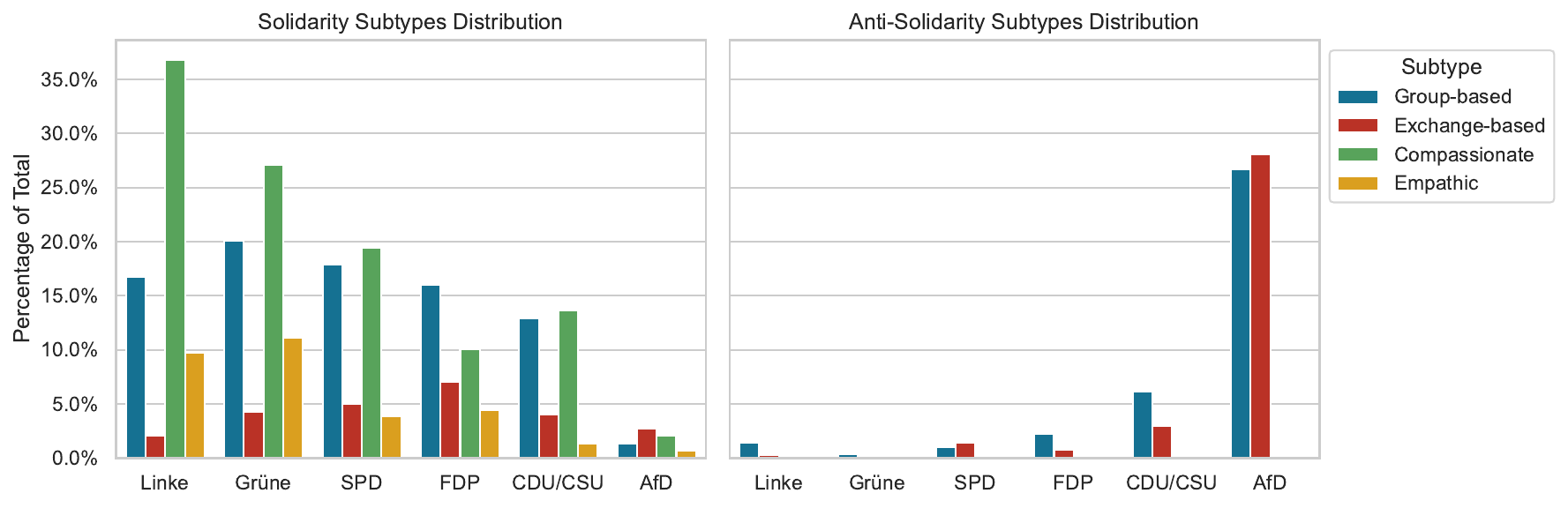}
    \caption{Distribution of \textit{(Anti-)solidarity} subtypes across selected political parties, ordered from the most left-wing to the most right-wing. Each subtype's percentage represents its share of the total statements for the corresponding party. \textit{Mixed} and \textit{None} categories are included in the distribution calculations but are not displayed in the visualization.}
    \figurelabel{fig:party-subtypes}
\end{figure*}

\begin{figure*}[!htb]
    \centering
    \includegraphics[width=0.5\linewidth]{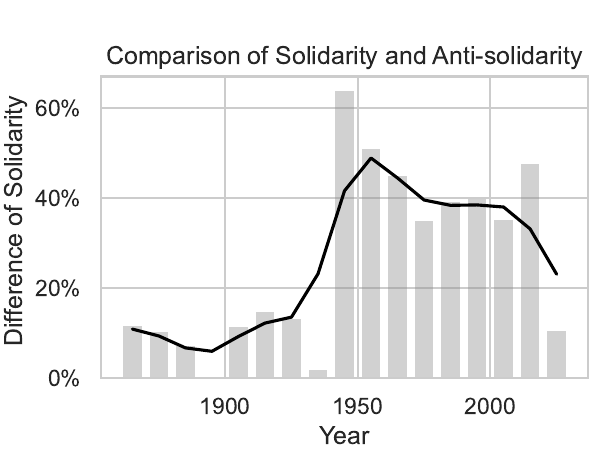}
    \caption{The percentage of solidarity instances minus the percentage of anti-solidarity instances, i.e., $\frac{\text{solidarity} - \text{anti-solidarity}}{\text{solidarity} + \text{anti-solidarity} + \text{mixed} + \text{none}}$, where negative values indicate more anti-solidarity; positive values more solidarity.}
    \figurelabel{fig:comparison-per-decade}
\end{figure*}

\begin{figure*}[!htb]
    \centering
    \includegraphics[width=\linewidth]{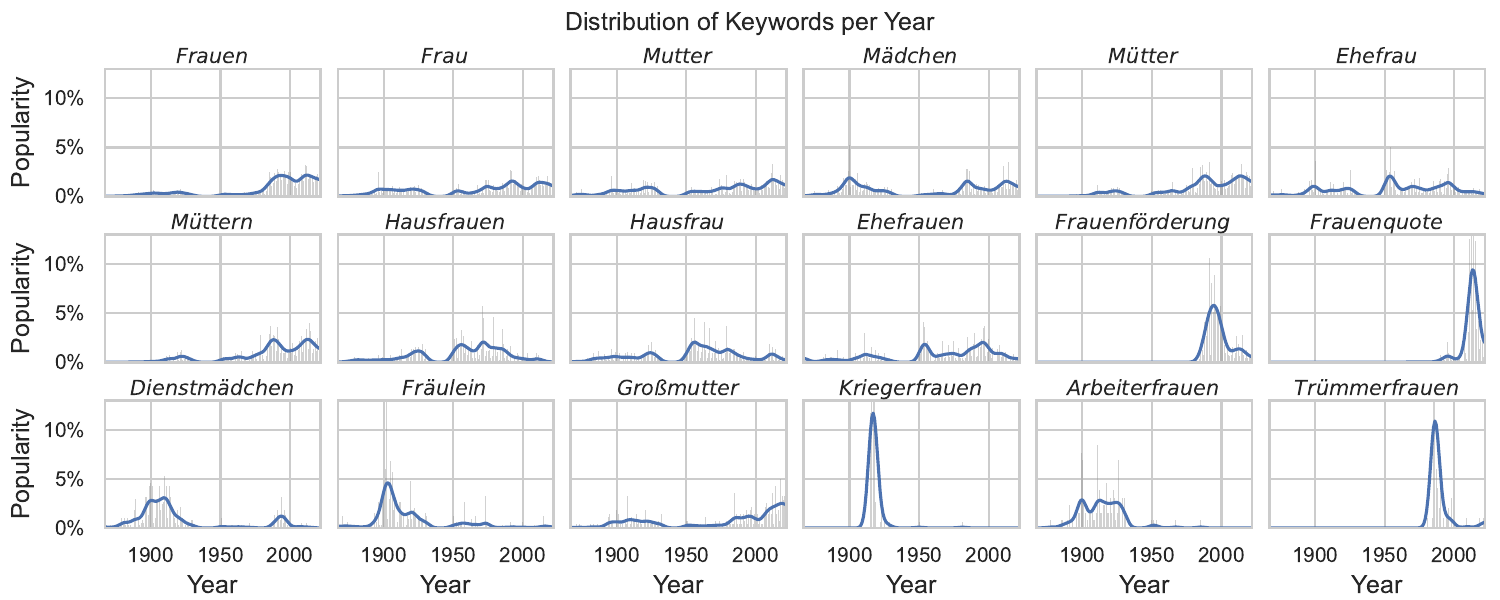}
    \caption{Distribution of all \Frau{} keywords over the years, normalized per keyword. The keywords are sorted by frequency, which means that the reliability decreases towards the bottom-right.}
    \figurelabel{fig:frau-keywords-per-year}
\end{figure*}

\begin{figure*}[!htb]
    \centering
    \includegraphics[width=\linewidth]{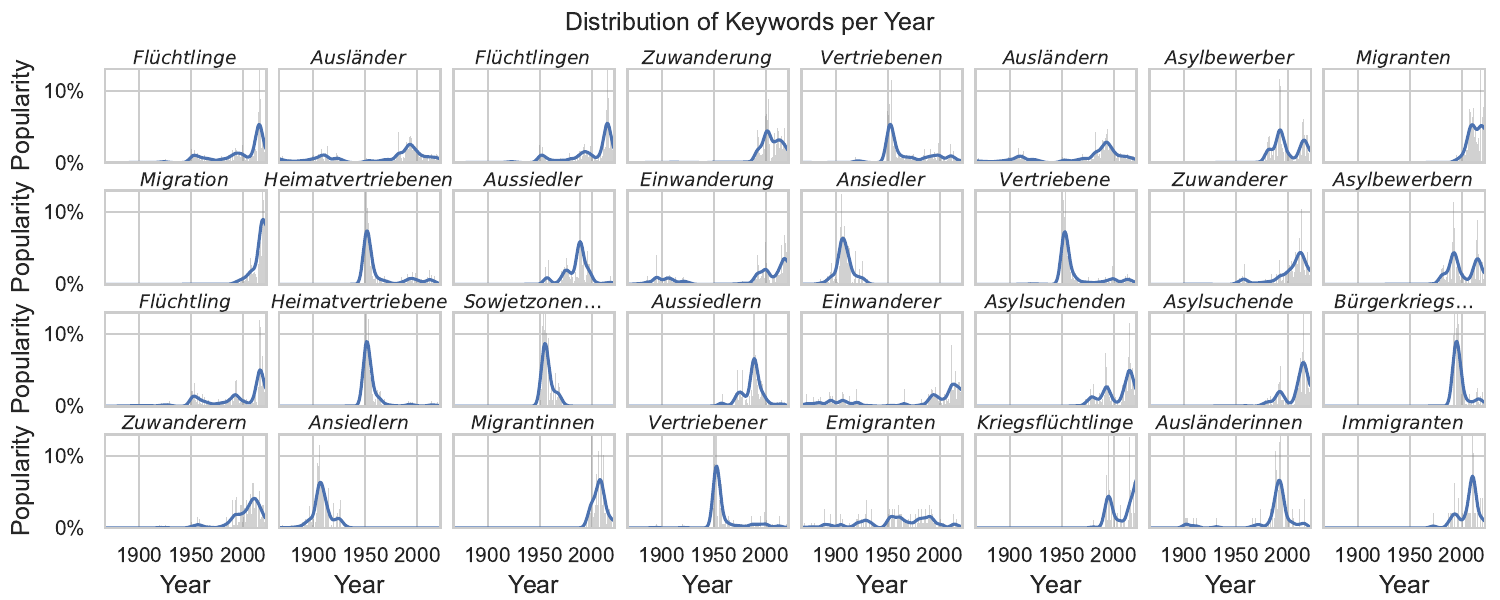}
    \caption{Distribution of all \Migrant{} keywords over the years, normalized per keyword. The keywords are sorted by frequency, which means that the reliability decreases towards the bottom-right.}
    \figurelabel{fig:migrant-keywords-per-year}
\end{figure*}

\begin{figure*}[!htb]
    \centering
    \includegraphics[width=\linewidth]{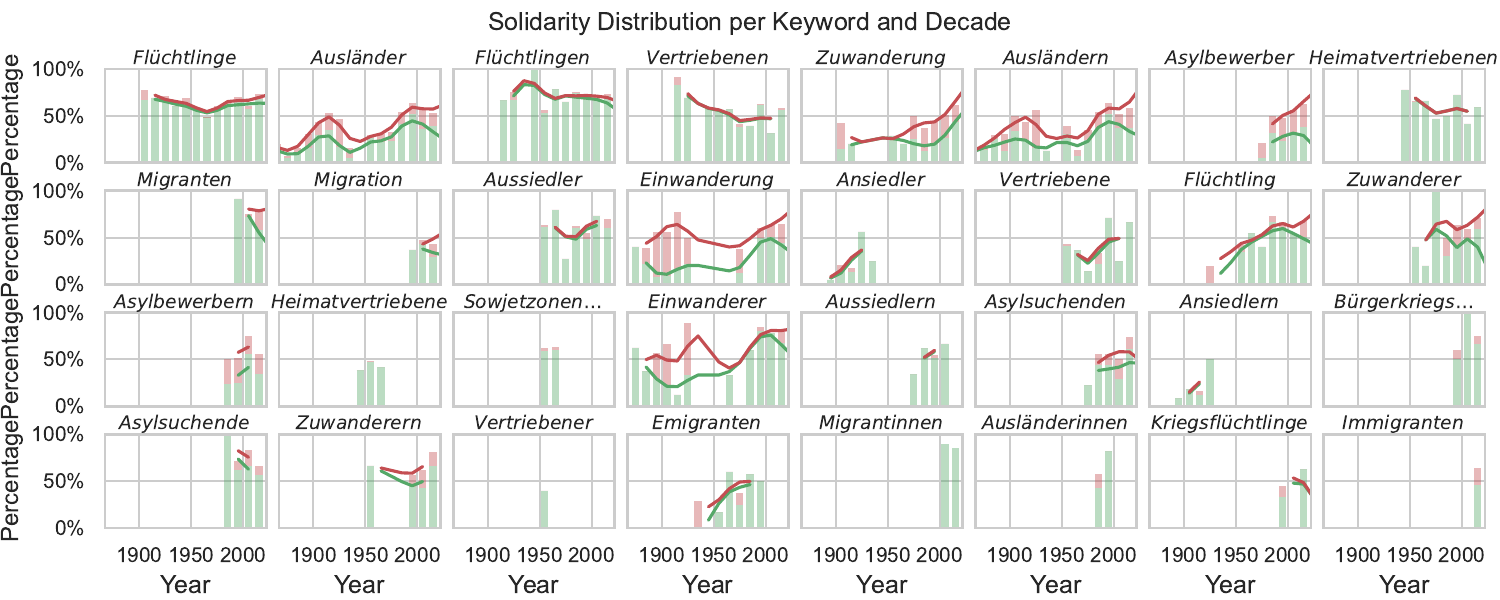}
    \caption{Percentage of sentences showing solidarity/anti-solidarity per decade for all \Migrant{} keywords. The keywords are sorted by frequency, which means that the reliability decreases towards the bottom-right.}
    \figurelabel{fig:solidarity-per-keyword-and-decade-migrant}
\end{figure*}

\begin{table*}[t]
    \centering
    {\footnotesize
    \begin{tabularx}{\textwidth}{XX}
        \toprule
        \textbf{Text} & \textbf{Gold Standard, Alternative Label \& Explanation} \\
        \midrule
        \enquote{[...] wenn nachher irgendwelche Schwierigkeiten bei der Rückzahlung der Darlehne entstehen, man nicht nach der Strenge des Gesetzes auf dem Schein bestehen und die Rückzahlung unter allen Umständen fordern müsse. \textbf{Ich habe aber noch eins ganz kurz zu bemerken: die Ausländer sollen auch mitberücksichtigt werden; nach unsern Beschlüssen in der Kommission würden die Ausländer in derselben Weise behandelt werden.} Auch da würde man erst fragen: bist du arm und hilfsbedürftig geworden, dann bekommst du eine Beihilfe als Geschenk; willst du dich weiter ansiedeln im Lande, dann bekommst du ein Darlehn. Ich möchte hier wiederholen: wir haben bei den deutschen Reichsangehörigen, die in andern Ländern geschädigt worden sind durch Revolution, uns nicht damit begnügt, daß sie eine Unterstützung bekommen haben, sondern wir haben erklärt: der Mann ist geschädigt, und er muß daher eine Entschädigung für seine Verluste erhalten, wenn er seiner Pflicht strengster Neutralität genügt hat, und diese Ansprüche haben wir nicht nur für Hilfsbedürftige erhoben, sondern auch für recht reiche Leute. [...]}
        & \textbf{Compassionate solidarity} or group-based solidarity towards migrants\hfill\break(Apr. 22, 1904)\hfill\break\hfill\break
        On the one hand, this text expresses compassionate solidarity by offering assistance to those in need, such as financial aid and loans, without requiring reciprocation. On the other hand, the speaker proposes treating foreigners and nationals equally in hardship, expressing group-based solidarity through a unified approach to support. \\
        \midrule
        \enquote{Dann bitte ich um genaue Nennung. Ich finde es in der Tat nicht sonderlich sinnvoll, daß wir amtliche deutsche Dokumente in einer nichtamtlichen Sprache abfassen. Wenn das geschehen ist, werden wir gerne darauf hinwirken, daß das geändert wird. \textbf{Ich möchte nur darauf aufmerksam machen, daß aus Ihrer Frage der völlig gegenteilige Sinn herauszulesen war, nämlich daß Sie offensichtlich wünschten, Ausländer, die in Deutschland Examen ablegen, sollten diese Diplome in ihrer eigenen Sprache ausgefertigt bekommen, was sicherlich nicht unsere Aufgabe sein kann und auch nicht sehr sinnvoll wäre.} [...]}
        & \textbf{Empathic anti-solidarity}, compassionate anti-solidarity or none towards migrants\hfill\break(Sept. 26, 1974)\hfill\break\hfill\break
        The text suggests empathic anti-solidarity by expecting foreign students to conform to German norms and shows compassionate anti-solidarity by deeming diplomas in native languages unnecessary. However, its focus on administrative details without strong bias classifies it as none. \\
        \midrule
        \enquote{Die Zieglerarbeit ist eine schwere, sogar eine sehr schwere; das wird allgemein anerkannt, auch von allen Gewerbeinspektoren. Es wird deshalb auch ziemlich häufig für wünschenswertst erklärt, daß die Arbeiterinnen aus diesem Produftionszweige mehr und mehr verdrängt werden. [...] \textbf{Wir sind sicherlich der Ansicht, daß die Arbeit auf den Ziegeleien im allgemeinen für Frauen nicht geeignet ist.} Deshalb können wir uns auch durchaus damit einverstanden erklären, daß man die Arbeit der Frauen auf den Ziegeleien erheblich eingeschränkt hat. Wir wünschen, daß man darin weiter fortfährt, auch selbst dann, wenn dadurch vielleicht zunächst ein gewisser Widerstand nicht nur bei den Unternehmern, sondern auch sogar bet den Arbeitern selber erzeugt werden wird. Denn darüber sind die Berichte ziemlich einig, daß, wenn die Frauenarbeit nicht erheblich eingeschränkt würde, dann die Frauen gesundheitliche und sittliche Schädigungen davontragen.}
        & \textbf{Empathic anti-solidarity} or mixed stance towards women\hfill\break(Jan. 13, 1897)\hfill\break\hfill\break
        The text can be classified as empathic anti-solidarity by suggesting that women be excluded from brickmaking, which supports traditional roles that limit their opportunities. It also presents a mixed stance by recognizing the job's difficulty and proposing to restrict women’s employment for their protection, which can be viewed as conditional support and simultaneously -- an imposed restriction. \\
        \midrule
        \enquote{Wir wünschen das nicht im Interesse der Frauen, wir haben uns mit den Frauenrechtlerinnen noch niemals auf eine Stufe gestellt. \textbf{Wir wünschen das nicht im Namen der Frauen und tm Interesse der Frauen, sondern im Gesamtinteresse des deutschen Volkes, weil wir der Meinung sind, daß bei Mitwirkung der Frauen mehr Verständnis für die Angeklagten und ein sozialerer Geist sich in der Rechtsprechung durchsetzen wird. Deshalb bitten wir Sie, bei den Beschlüssen zweiter Lesung zu bleiben.} [...]}
        & \textbf{Group-based solidarity} or exchange-based solidarity towards women\hfill\break(March 8, 1921)\hfill\break\hfill\break
        This text shows group-based solidarity by promoting women's participation in the judiciary to improve legal proceedings for societal benefit. It can also be interpreted as exchange-based solidarity, as it highlights the reciprocal advantages of women's inclusion. \\
        \bottomrule
    \end{tabularx}}
    \caption{Examples of divergence between our annotators. We mark the gold label bold and add explanations of why two or more labels could be correct, to illustrate the difficulty of this task. Bold text is the main sentence, the other sentences are for context.}
    \tablelabel{tab:annotator-divergence-examples}
\end{table*}

\begin{figure*}[!htb]
    \centering
    \subfloat[Annotators' agreement per decade.]{
        \includegraphics[width=0.5\linewidth]{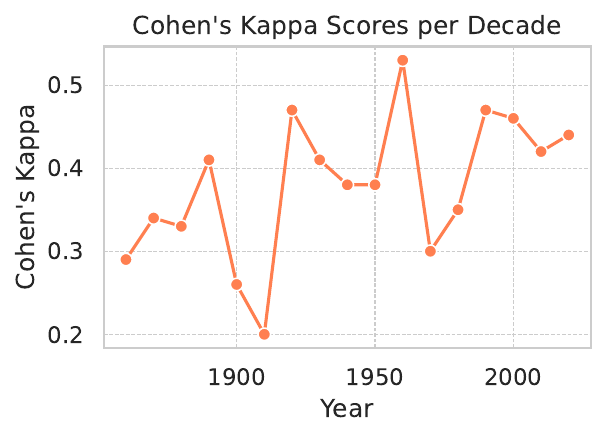}
        \label{subfig:kappa-over-time}
    }
    \subfloat[Macro F1 scores from GPT-4 over time.]{
        \includegraphics[width=0.5\linewidth]{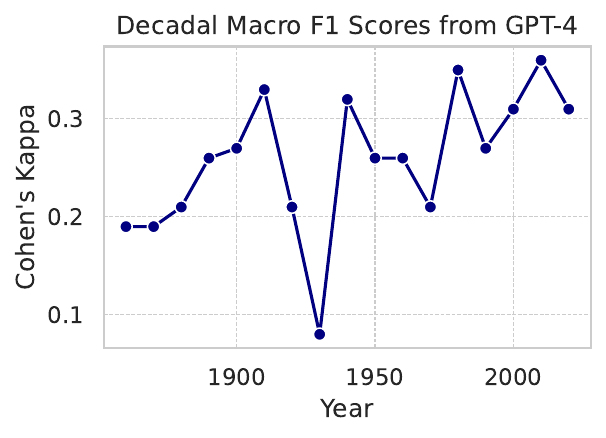}
        \label{subfig:f1-over-time}
    }
    \vspace{-0.1cm}
    \caption{Pairwise Cohen's Kappa agreement among human annotators and macro F1 scores of GPT-4, per decade.}
    \figurelabel{fig:agreement-f1-overtime}
\end{figure*}

\begin{table*}[htb!]
    \centering
    \renewcommand{\arraystretch}{1.2}
    \footnotesize
    \begin{tabular}{llcc|cccccccc}
        \toprule
        & & \multicolumn{2}{c}{Fine-grained (high-level)} & \multicolumn{2}{c}{Solidarity} & \multicolumn{2}{c}{Anti-solidarity} & \multicolumn{2}{c}{Mixed} & \multicolumn{2}{c}{None} \\
        \cmidrule(lr){3-4} \cmidrule(lr){5-6} \cmidrule(lr){7-8} \cmidrule(lr){9-10} \cmidrule(lr){11-12}
        Model & Method & W & M & W & M & W & M & W & M & W & M \\
        \midrule\midrule
        \multirow{2}{*}{\texttt{GPT-4}} 
        & 0-shot & \shortstack{\textbf{0.37 (0.60)}} & \shortstack{\textbf{0.42 (0.73)}} & \textbf{0.85} & \textbf{0.86} & \textbf{0.65} & \textbf{0.87} & 0.30 & \textbf{0.58} & \textbf{0.62} & 0.61 \\
        & Few-shot & \shortstack{\textbf{0.37 (0.54)}} & \shortstack{\textbf{0.43 (0.63)}} & \textbf{0.85} & 0.83 & 0.50 & 0.75 & 0.18 & 0.40 & 0.63 & 0.53 \\
        \midrule
        \multirow{2}{*}{\shortstack[l]{\texttt{GPT-3.5}\\\texttt{fine-tuned}}} 
        & 0-shot & 0.18 (0.45) & 0.27 (0.53) & 0.80 & 0.74 & 0.12 & 0.61 & 0.28 & 0.27 & 0.59 & 0.51 \\
        & Few-shot & 0.22 (0.47) & 0.28 (0.48) & 0.78 & 0.70 & 0.18 & 0.65 & 0.35 & 0.07 & 0.58 & 0.48 \\
        \midrule
        \multirow{2}{*}{\shortstack[l]{\texttt{GPT-3.5} \\ \texttt{base}}} 
        & 0-shot & \shortstack{0.15 (0.46)} & \shortstack{0.19 (0.48)} & 0.75 & 0.66 & 0.38 & 0.68 & 0.17 & 0.11 & 0.54 & 0.48 \\
        & Few-shot & \shortstack{0.12 (0.41)} & \shortstack{0.27 (0.50)} & 0.70 & 0.61 & 0.33 & 0.64 & 0.25 & 0.23 & 0.36 & 0.54 \\
        \midrule
        \multicolumn{2}{c}{\texttt{Llama3-70B 0-shot}} & \shortstack{0.24 (0.48)} & \shortstack{0.27 (0.65)} & 0.77 & 0.79 & 0.41 & 0.78 & \textbf{0.36} & 0.31 & 0.39 & \textbf{0.73} \\
        \midrule
        \multicolumn{2}{c}{\texttt{BERT}} & \shortstack{0.13 (0.26)} & \shortstack{0.24 (0.46)} & 0.75 & 0.74 & 0.00 & 0.51 & 0.08 & 0.10 & 0.20 & 0.51 \\
        \midrule
        \midrule
        \multicolumn{2}{c}{Human upper bound} & \shortstack{0.48 (0.72)} & \shortstack{0.56 (0.78)} & 0.87 & 0.88 & 0.68 & 0.86 & 0.57 & 0.64 & 0.76 & 0.74 \\
        \bottomrule
    \end{tabular}
    \caption{Comparative performance (macro F1) of models vs. human upper bound (calculated as an average macro F1 between annotators’ labels and the final label) on combined high-level (in parentheses) and fine-grained tasks for both women (W) and migrants (M), with further F1 scores for the categories of solidarity, anti-solidarity, and mixed stance. Best scores for fine-grained tasks and per label are highlighted in bold.}
    \label{tab:models_performance}
\end{table*}

\begin{table*}[t]
    \centering
    {\footnotesize
    \begin{tabularx}{\textwidth}{>{\hsize=.5\hsize\raggedright\arraybackslash}X >{\hsize=1.4\hsize}X >{\hsize=1.1\hsize}X}
        \toprule
        \textbf{} & \textbf{Original Text} & \textbf{Translation} \\
        \midrule
        (1) \textbf{Gold standard: \newline{}group-based solidarity \newline{}towards migrants\newline{}\newline{} Predicted label: none}\newline{}(Sept. 14, 1989)
        & \enquote{[...] \textbf{Dr. Hirsch: Herr Minister Möllemann, es ist doch wohl so, daß es sich nicht um das Geld des Bundes oder um das Geld eines Landes handelt, sondern immer um das Geld des Steuerzahlers und daß dementsprechend verlangt werden kann, daß Bund und Länder gemeinsam das tun, was im Interesse der Bevölkerung, zu der dann ja auch die Aussiedler gehören, notwendig und richtig ist.} [...]}
        & \enquote{[...] \textbf{Dr. Hirsch: Mr. Minister Möllemann, it is indeed the case that it is not about the federal government's money or a state's money, but always about the taxpayer's money, and accordingly, it can be demanded that the federal and state governments together do what is necessary and right in the interest of the population, which then also includes the expellees}. [...]} \\
        \midrule
        \textbf{Model's Explanation} & \multicolumn{2}{>{\hsize=2.4\hsize}X}{\textit{[...] The appropriate high-level category for this text is NONE, as it neither promotes support nor opposition towards migrants but rather discusses financial governance regarding a subset of migrants without an emotional or ideological angle.}} \\
        \midrule
        (2) \textbf{Gold standard: \newline{}exchange-based solidarity \newline{}towards migrants\newline{}\newline{} Predicted label: mixed stance}\newline{}(May 26, 1869)
        & \enquote{[...] \textbf{Ich glaube wohl nicht näher aus die materiellen Unzuträglichkeiten eingehen zu sollen, die damit verbunden sind, wenn man die Gegenseitigkeit in der Gesetzgebung dem Ausländer gegenüber streng ausrecht erhält.} Es liegt ja doch wirklich im Interesse der Staaten des Norddeutschen Bundes, daß man diejenigen Ausländer, die uns etwas Gutes bringen können, zum Gewerbebetrieb zuläßt und uns nicht selbst damit straft, daß man die Gewerbetreibenden nicht zuläßt, in deren Heimath vielleicht noch sehr schlechte Gewerbegesetze existiren. [...]}
        & \enquote{[...] \textbf{I believe I should not delve into the material inconveniences associated with maintaining strict reciprocity in legislation towards foreigners.} It is indeed in the interest of the states of the North German Confederation to allow those foreigners who can bring us benefits to engage in business, and not to punish ourselves by denying business people who may come from countries with very poor trade laws. [...]} \\
        \midrule
        \textbf{Model's Explanation} & \multicolumn{2}{>{\hsize=2.4\hsize}X}{\textit{[...] This could be interpreted as exchange-based solidarity to some extent. However, the core of the discussion is legal and pragmatic, rather than being based on an ethos of solidarity or anti-solidarity as defined. [...] It is important to note that without the broader context of the discussion or the surrounding text, it is difficult to ascertain the overall stance regarding migrants beyond this specific legislative detail.}} \\
        \midrule
        (3) \textbf{Gold standard: \newline{}compassionate solidarity \newline{}towards women\newline{}\newline{} Predicted label: group-based solidarity}\newline{}(March 13, 2003)
        & \enquote{[...] \textbf{Wir werden es nicht zulassen, dass künftig ausländische Ehefrauen vier Jahre in unzumutbaren Ehen verharren, um nicht aus Deutschland ausgewiesen zu werden.} Das werden Sie nicht erreichen. Ich will es überhaupt nicht verhehlen: Wir haben noch unendlich viel zu tun, bevor wir gerade im Erwerbsleben und bei der traditionellen Rollen- und Aufgabenverteilung eine faktische Gleichstellung von Frauen und Männern erreicht haben. [...]}
        & \enquote{[...] \textbf{We will not allow foreign wives to remain in unbearable marriages for four years to avoid being deported from Germany}. You will not achieve this. I do not want to conceal it at all: we still have an immense amount of work to do before we achieve actual equality between women and men in the workforce and in the traditional distribution of roles and tasks. We take this responsibility seriously. [...]} \\
        \midrule
        \textbf{Model's Explanation} & \multicolumn{2}{>{\hsize=2.4\hsize}X}{\textit{[...] The most fitting subtype of solidarity in this context is GROUP-BASED SOLIDARITY because the text discusses the collective responsibility to promote equality and the shared efforts needed to achieve substantial change in societal structures concerning the roles and treatment of women.}} \\
        \midrule
        (4) \textbf{Gold standard: \newline{}compassionate solidarity \newline{}towards women\newline{}\newline{} Predicted label: empathic solidarity}\newline{}(Apr. 5, 1922)
        & \enquote{[...] \textbf{Wenn man von den Leuten fordert, daß sie sich anmelden sollen, wenn alle möglichen Einschränkungen für den Aufenthalt getroffen werden, soll man diese Ausländer auch wissen lassen, nach welchen Vorschriften sie sich zu richten haben. Es ist schikanös und heimtückisch, die Leute mit solchen Vorschriften zu überfallen.} Wie das in der Praxis vor sich geht, brauche ich Ihnen nicht lange zu schildern. [...]}
        & \enquote{[...] \textbf{When people are required to register, and all kinds of restrictions on their stay are imposed, these foreigners should also be informed about the regulations they must follow.} It is harassing and treacherous to ambush people with such regulations. I do not need to describe to you at length how this happens in practice. [...]} \\
        \midrule
        \textbf{Model's Explanation} & \multicolumn{2}{>{\hsize=2.4\hsize}X}{\textit{[...] The text seems to transmit a sense of EMPATHIC SOLIDARITY because it addresses the unfairness and challenges that foreigners face and advocates for fairness and clear guidelines, which shows a degree of empathy for their situation.}} \\
        \bottomrule
    \end{tabularx}}
    \caption{Example predictions by GPT-4 for both target groups in 0-shot setting for migrants and women (solidarity cases). Bold text is the main sentence, the other sentences are for context.}
    \tablelabel{tab:predictions-examples}
\end{table*}

\begin{table*}[t]
    \centering
    {\footnotesize
    \begin{tabularx}{\textwidth}{>{\hsize=.5\hsize\raggedright\arraybackslash}X >{\hsize=1.4\hsize}X >{\hsize=1.1\hsize}X}
        \toprule
        \textbf{} & \textbf{Original Text} & \textbf{Translation} \\
        \midrule
        (5) \textbf{Gold standard: \newline{}group-based anti-solidarity \newline{}towards migrants\newline{}\newline{} Predicted label: exchange-based anti-solidarity}\newline{}(Apr. 29, 1921)
        & \enquote{[...] \textbf{Wir verlangen des weiteren aber auch schärfste Kontrolle der Ausländer.} Bei dieser Gelegenheit möchte ich an die Retchsregierung die Frage richten, ob es richtig ist, daß vom Auswärtigen Amt zehn Russen das Recht erhalten haben, Einreiseerlaubnisscheine nach Deutschland auszustellen. Wir verlangen schärfste Kontrolle der Ausländer. Nach dem, was ich aber eben mitgeteilt habe, würde natürlich eine Kontrolle illusorisch sein.}
        & \enquote{[...] \textbf{We also demand the strictest control of foreigners.}. On this occasion, I would like to ask the Reich government whether it is true that the Foreign Office has granted ten Russians the right to issue entry permit certificates to Germany. We demand the strictest control of foreigners. However, based on what I have just reported, such control would naturally be illusory.} \\
        \midrule
        \textbf{Model's Explanation} & \multicolumn{2}{>{\hsize=2.4\hsize}X}{\textit{[...] Given the nature of the concerns highlighted, which are about resource distribution (control and monitoring), the sentiment somewhat aligns with an exchange-based view but from a critical angle, focusing on preventing perceived negative impacts rather than fostering an exchange. Therefore, the most appropriate classification would be EXCHANGE-BASED ANTI-SOLIDARITY.}} \\
        \midrule
        (6) \textbf{Gold standard: \newline{}exchange-based anti-solidarity \newline{}towards migrants\newline{}\newline{} Predicted label: none}\newline{}(May 14, 1982)
        & \enquote{[...] Der Flughafen Frankfurt ist Einfallstor für Asylbewerber. \textbf{Das Land Hessen ist daher von dem Zustrom der Asylbewerber besonders betroffen.} Obgleich die Zahl der Asylbewerber seit 1980 rückläufig ist, hat Hessen immer noch einen Überhang von 1 500 Asylbewerbern. Im Jahr 1981 hat das Land allein auf Grund seines Aufnahmegesetzes 22,4 Millionen DM für die Asylbewerber aufzubringen gehabt.}
        & \enquote{[...] Frankfurt Airport is a gateway for asylum seekers. \textbf{Therefore, the state of Hesse is particularly affected by the influx of asylum seekers.} Although the number of asylum seekers has been declining since 1980, Hesse still has a surplus of 1,500 asylum seekers. In the year 1981, the state had to raise 22.4 million DM for asylum seekers alone, based on its reception law.} \\
        \midrule
        \textbf{Model's Explanation} & \multicolumn{2}{>{\hsize=2.4\hsize}X}{\textit{[...] the text does not display solidarity with or anti-solidarity against asylum seekers in the context of promoter solidarity or anti-solidarity subtypes as defined. It seems to be merely stating the situation as a matter of fact without expressing a clear stance towards the asylum seekers.}} \\
        \midrule
        (7) \textbf{Gold standard: \newline{}empathic anti-solidarity \newline{}towards women\newline{}\newline{} Predicted label: none}\newline{}(April 25, 1974)
        & \enquote{[...] Eben ist von Frau Kollegin Funcke gesagt worden, das Fristenmodell sei einfach und klar. \textbf{Einfach und klar für die Frau, aber auch einfach und klar für das ungeborene Leben: es wird einfach und klar getötet.} Die von uns vorgeschlagene Lösung unterscheidet sich vom Fristenmodell prinzipiell dadurch, daß wir sagen: Erstens. In einer Konfliktsituation zwischen Frau und ungeborenem Kind kann nicht die Frau allein über das ungeborene Leben verfügen. Wir legen diese Entscheidung in die Hände von Ärzten.}
        & \enquote{[...] Just now, colleague Funcke said that the deadline model is simple and clear. \textbf{Simple and clear for the woman, but also simple and clear for the unborn life: it is simply and clearly killed.} The solution we propose differs fundamentally from the deadline model in that we say: First. In a conflict situation between a woman and an unborn child, the woman alone cannot decide over the unborn life. We place this decision in the hands of doctors.} \\
        \midrule
        \textbf{Model's Explanation} & \multicolumn{2}{>{\hsize=2.4\hsize}X}{\textit{[...] Given the text's focus on an ethical debate that does not inherently align with either supporting or opposing women's societal standing or collective rights, the most fitting classification appears to be NONE.}} \\
        \midrule
        (8) \textbf{Gold standard: \newline{}compassionate anti-solidarity \newline{}towards women\newline{}\newline{} Predicted label: exchange-based solidarity}\newline{}(March 14, 2013)
        & \enquote{Aber Sie gehören nach wie vor einer Regierung an, die [...] ein idiotisches Betreuungsgeld ausreicht mit allen Folgen, die das für den Arbeitsmarkt hat. Wenn wir das Thema der Fachkräftesicherung ernst nehmen, dann müssen wir dafür sorgen, dass vor allen Dingen Frauenerwerbsbeteiligung in Vollzeit in diesem Land zum Zuge kommt. [...] \textbf{Aber es gilt auch in diesem Fall, Frau von der Leyen: Nicht reden, sondern handeln.} [...]}
        & \enquote{But you still belong to a government that [...] provides an idiotic childcare allowance, with all the consequences that has for the labor market. If we take the issue of securing skilled labor seriously, then we must ensure that, above all, women's participation in full-time employment in this country comes into effect. [...] \textbf{But in this case, too, Mrs. von der Leyen: not talk, but act.} [...]} \\
        \midrule
        \textbf{Model's Explanation} & \multicolumn{2}{>{\hsize=2.4\hsize}X}{\textit{[...] The text is critical of a policy deemed detrimental to promoting women's full participation in the workforce and calls for action rather than mere discussion. The critique is embedded within an appeal to recognize the importance of women's employment in addressing broader societal and economic challenges. [...]}} \\
        \bottomrule
    \end{tabularx}}
    \caption{Example predictions by GPT-4 for both target groups in 0-shot setting for migrants and women (anti-solidarity cases). Bold text is the main sentence, the other sentences are for context.}
    \tablelabel{tab:predictions-examples}
\end{table*}

\begin{figure*}[!htb]
    \centering
    \includegraphics[width=\linewidth]{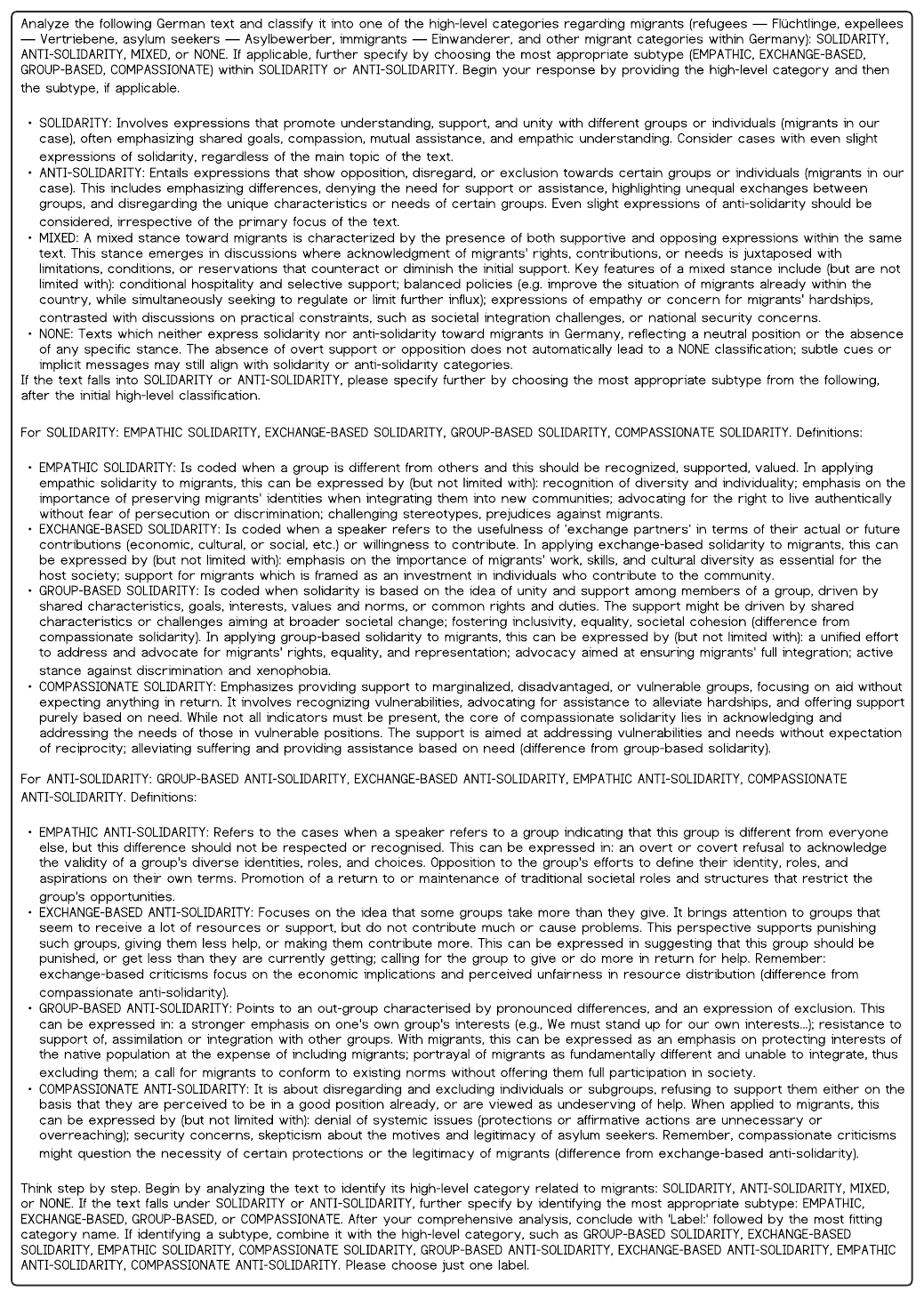}
    \caption{2-step Prompt for Migrants used for GPT 0-shot experiments}
    \figurelabel{fig:prompt-migrant-appendix}
\end{figure*}

\begin{figure*}[!htb]
    \centering
    \includegraphics[width=\linewidth]{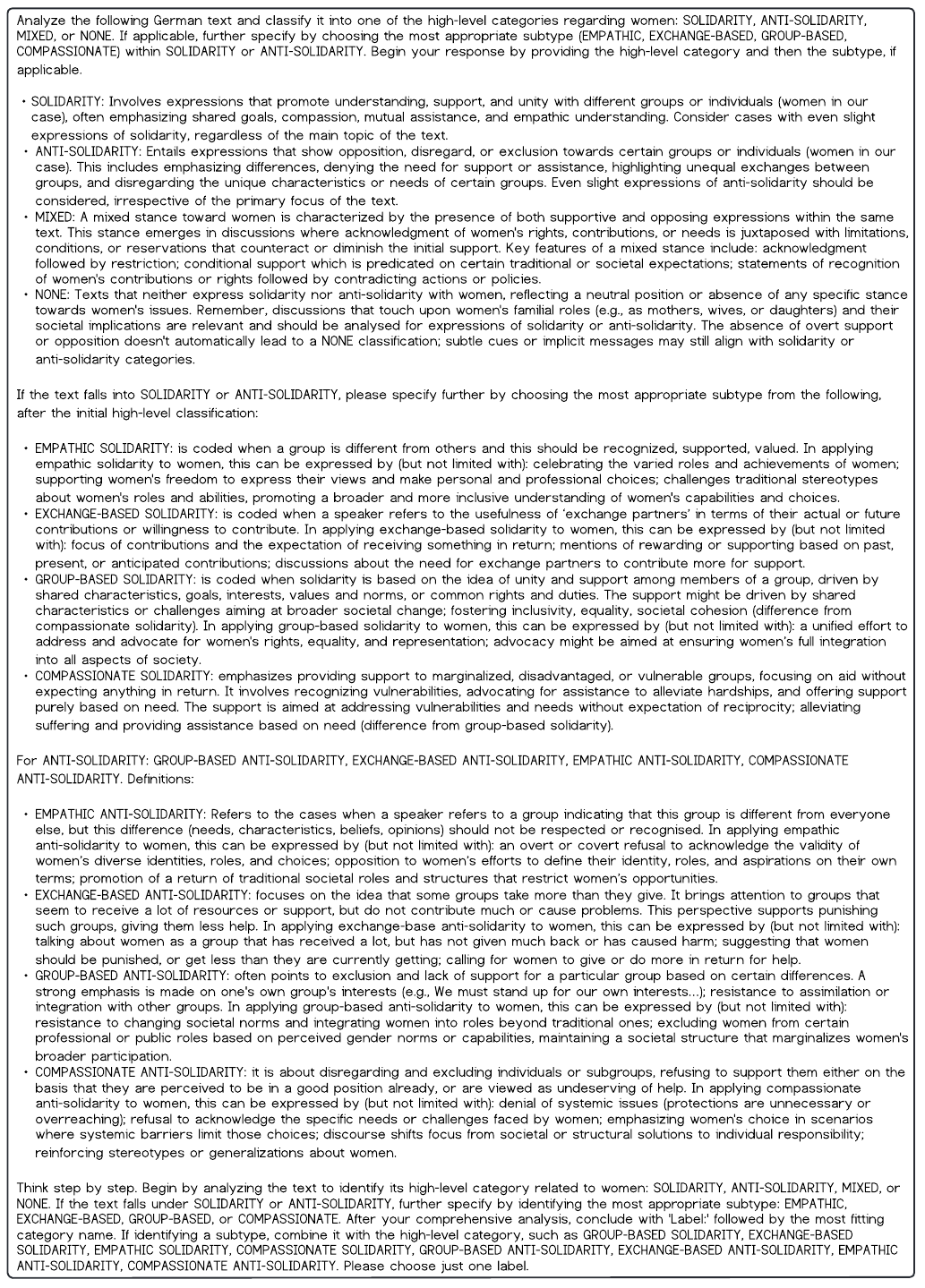}
    \caption{2-step Prompt for Women used for GPT 0-shot experiments}
    \figurelabel{fig:prompt-frau-appendix}
\end{figure*}

\end{document}